\documentclass[10pt,twocolumn,letterpaper]{article}

\usepackage[pagenumbers]{cvpr} 
\usepackage{url}
\usepackage{multirow}
\usepackage{bm}
\usepackage{booktabs}
\usepackage{amsmath}
\usepackage{amssymb}
\usepackage{graphicx}

\usepackage[pagebackref=true,breaklinks=true,colorlinks,bookmarks=false,citecolor=blue,linkcolor=blue,urlcolor=blue]{hyperref}

\begin{document}
\title{Gated Multi-Resolution Transfer Network for Burst Restoration and Enhancement}

\author{Nancy Mehta$^1$ \quad Akshay Dudhane$^2$ \quad Subrahmanyam Murala$^{1}$ \quad Syed Waqas Zamir$^3$ \\ Salman Khan$^{2,4}$ \quad Fahad Shahbaz Khan$^{2,5}$ \\
$^1$CVPR Lab, Indian Institute of Technology Ropar \hspace{1.5mm} $^2$Mohamed bin Zayed University of AI \hspace{1.5mm}  \\
$^3$Inception Institute of AI \hspace{1.5mm} $^4$Australian National University
\hspace{1.5mm} $^5$Link\"{o}ping University
 }
\maketitle

\begin{abstract}
    Burst image processing is becoming increasingly popular in recent years.
    However, it is a challenging task since individual burst images undergo multiple degradations and often have mutual misalignments resulting in ghosting and zipper artifacts.
    Existing burst restoration methods
    usually do not consider the mutual correlation and non-local contextual information among burst frames, which tends to limit these approaches in challenging cases.
    Another key challenge lies in the robust up-sampling of burst frames. The existing up-sampling methods cannot effectively utilize the advantages of single-stage and progressive up-sampling strategies with conventional and/or recent up-samplers at the same time.
    To address these challenges, we propose a novel \textbf{G}ated \textbf{M}ulti-Resolution \textbf{T}ransfer Network (GMTNet) to reconstruct a spatially precise high-quality image from a burst of low-quality raw images.
    GMTNet consists of three modules optimized for burst processing tasks: Multi-scale Burst Feature Alignment (MBFA) for feature denoising and alignment, Transposed-Attention Feature Merging (TAFM) for multi-frame feature aggregation, and Resolution Transfer Feature Up-sampler (RTFU) to up-scale merged features and construct a high-quality output image.
    Detailed experimental analysis on five datasets validates our approach and sets a  state-of-the-art for burst super-resolution, burst denoising, and low-light burst enhancement. Our codes and models are available at \url{https://github.com/nanmehta/GMTNet}.
\end{abstract}

\section{Introduction}
\label{sec:intro}
    \begin{figure}[t]
        \centering
        \includegraphics[width=0.95\linewidth]{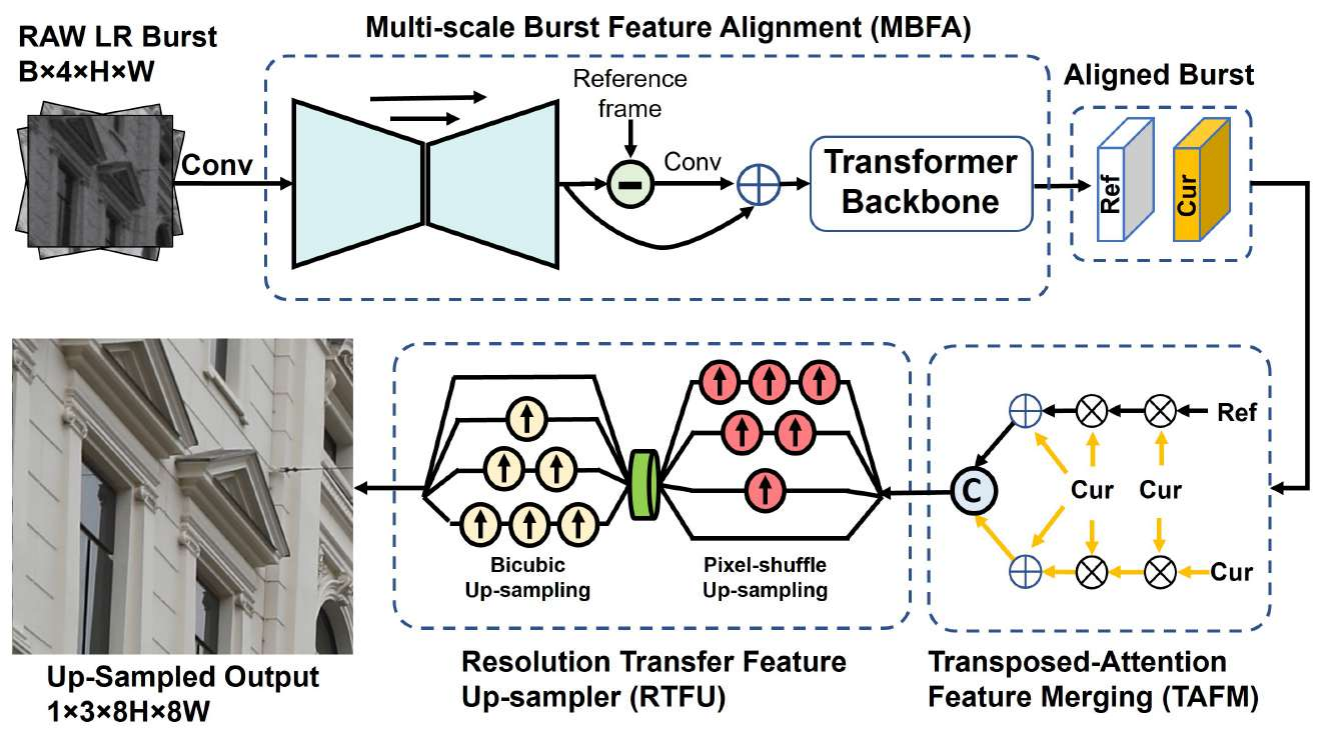}
        
        \caption{Proposed GMTNet processes RAW burst LR frames and gives a high-quality image through three key stages: (1) Multi-scale Burst Feature Alignment (MBFA), (2) Transposed-Attention Feature Merging (TAFM), and (3) Resolution Transfer Feature Up-sampler (RTFU).}
        \label{fig: 1}
    \end{figure}
    With the soaring popularity of smartphones in day-to-day life, the demand for capturing high-quality images is rapidly increasing. 
    However, the camera in the smartphone has several limitations due to the constraints placed on it to be integrated into its thin profile.
    The most prominent hardware limitations are the small camera sensor size and the associated lens optics that reduce their spatial resolution and dynamic range \cite{delbracio2021mobile}, impeding them in reconstructing DSLR-like images. 
    To deal with these inherent physical limitations of mobile photography, one emerging solution is to leverage multi-frame (burst) processing instead of single-frame processing.
    Burst processing techniques primarily focus on extracting high-frequency details by merging non-redundant data from various shifted images to produce a high-quality image.
    Three critical factors involved in burst processing are feature alignment, fusion, and subsequent reconstruction of the obtained frames. 
    Generally, any burst processing approach is limited by the accuracy of the alignment process on account of the camera and scene motion of dynamically moving objects.
    Therefore, it is crucial to design a module for facilitating accurate alignment, as the subsequent fusion and reconstruction modules must be robust to misalignment for generating an artifact-free image. 
    We further note that the alignment and fusion modules in existing burst processing approaches \cite{dudhane,bhat2} do not consider the non-local dependencies and mutual correlation among the frames which hinders the flexible inter-frame information exchange. 
    Moreover, the existing burst up-sampling approaches \cite{bhat1,dudhane}  do not take into account the merits of repeatedly transferring the information across several resolutions. 
    To address these issues, we present a novel burst processing framework named Gated Multi-Resolution Transfer network (GMTNet) as illustrated in Figure \ref{fig: 1}.

    In  contrast to the previous works \cite{bhat1, bhat2} which adopt bulky pre-trained modules for alignment, we propose an implicit Multi-scale Burst Feature Alignment (MBFA) to reduce the inter-frame misalignment.
    Overall, MBFA module implicitly learns feature alignment at multiple scales through the proposed Attention-Guided Deformable Alignment (AGDA) module and obtains an enriched feature representation via Aligned Feature Enrichment (AFE) module.
    The proposed AFE module is composed of a back-projection mechanism and capable of extracting long-range pixel interactions that ease the feature alignment in complex motions, where simply aligning the frames does not suffice.
    Additionally, unlike the recent state-of-the-art (SoTA) algorithm, BIPNet \cite{dudhane} that utilizes a computationally intensive pseudo burst mechanism on the aligned burst for inter-frame communication, we propose a simple Transposed-Attention based Feature Merging (TAFM) module that leverages local and non-local correlations to allow an extensive interaction with the reference frame. 
    Finally, our Resolution Transfer Feature Up-sampler (RTFU) combines the complementary features of both single-stage and progressive up-sampling strategies through deployed conventional and recent feature up-samplers.
    Such a design enables strong feature embedding of LR and HR images that creates a solid foundation for up-sampling in burst SR tasks. 
    \noindent In this work, we validate our GMTNet for popular burst processing tasks such as super-resolution, denoising and low-light image enhancement. Overall, the following are our key contributions.
    \begin{enumerate}\setlength{\itemsep}{0em}
       \item A Multi-scale Burst Feature Alignment (MBFA) is proposed which uses both local and non-local features for alignment at multiple scales, resolving the spatial misalignment within burst images (\textsection \ref{TPAM}). 
       \item A Transposed-Attention Feature Merging (TAFM) is proposed to aggregate the features of the aligned and reference frames (\textsection \ref{TAFM}). 
       \item A Resolution Transfer Feature Up-sampler (RTFU) is proposed to upscale the merged features. The proposed RTFU integrates the complementary features extracted by single-stage and progressive up-sampling strategies using the conventional and recent up-samplers (\textsection \ref{RTFU}). 
    \end{enumerate}
    Our three-stage design achieves SoTA results on both synthetic as well as real raw datasets for burst super-resolution, denoising and low-light enhancement.
    %
%
\section{Related Work}
\label{sec:II}
    \paragraph{Multi-Frame Super-Resolution.}
        Compared to the single-image super-resolution (SISR), multi-frame super-resolution (MFSR) encounters new challenges while estimating the offsets among different images caused by camera movement and moving objects.
        Tsai and Huang \cite{tsai} were the first to put forward a computationally cheap, frequency domain-based solution for the MFSR problem. 
        Due to significant visual artifacts in frequency domain processing, spatial domain algorithms gained popularity \cite{spatial,robust1}.      
        Following it, Irani and Peleg \cite{irani} and Peleg \textit{et al.} \cite{peleg} proposed an iterative back-projection-based approach, and \cite{map1} utilized maximum a posteriori (MAP) model to obtain better super-resolved results. 
        But all the above-mentioned approaches were based upon the assumption that motion between input frames, as well as the image formation model can be well estimated. 
        Subsequent works addressed this issue with the joint estimation of the unknown parameters \cite{u1, u2}.
        \par Recently, a few data-driven approaches have been proposed for different applications, such as satellite imaging \cite{deudon} and medical images \cite{medical}. %
        Bhat \textit{et al.} \cite{bhat1} addressed the problem of MFSR~\cite{bhat2021ntire, bhat2022ntire} by proposing an explicit feature alignment and attention-based fusion mechanism. 
        However, explicit motion estimation and image-warping techniques can pose difficulty in handling scenes with fast object motions. 
        Dudhane \textit{et al.} \cite{dudhane,new3} proposed a generalized approach for processing noisy raw bursts through their implicit feature alignment and inter-frame communication strategy.
        Despite its better accuracy, \cite{dudhane} fails to consider the relevant non-local contextual information at multiple scales while aligning and fusing the features.
        
        \vspace{0.4em}
        \noindent\textbf{Multi-frame Denoising.}
        Existing methods either utilize neural networks that are purely feed-forward \cite{feed1, feed2}, recurrent networks \cite{rec} or a hybrid of both \cite{both} for multi-frame denoising.
        \begin{figure*}[t]
            \centering
            \includegraphics[width=1\linewidth]{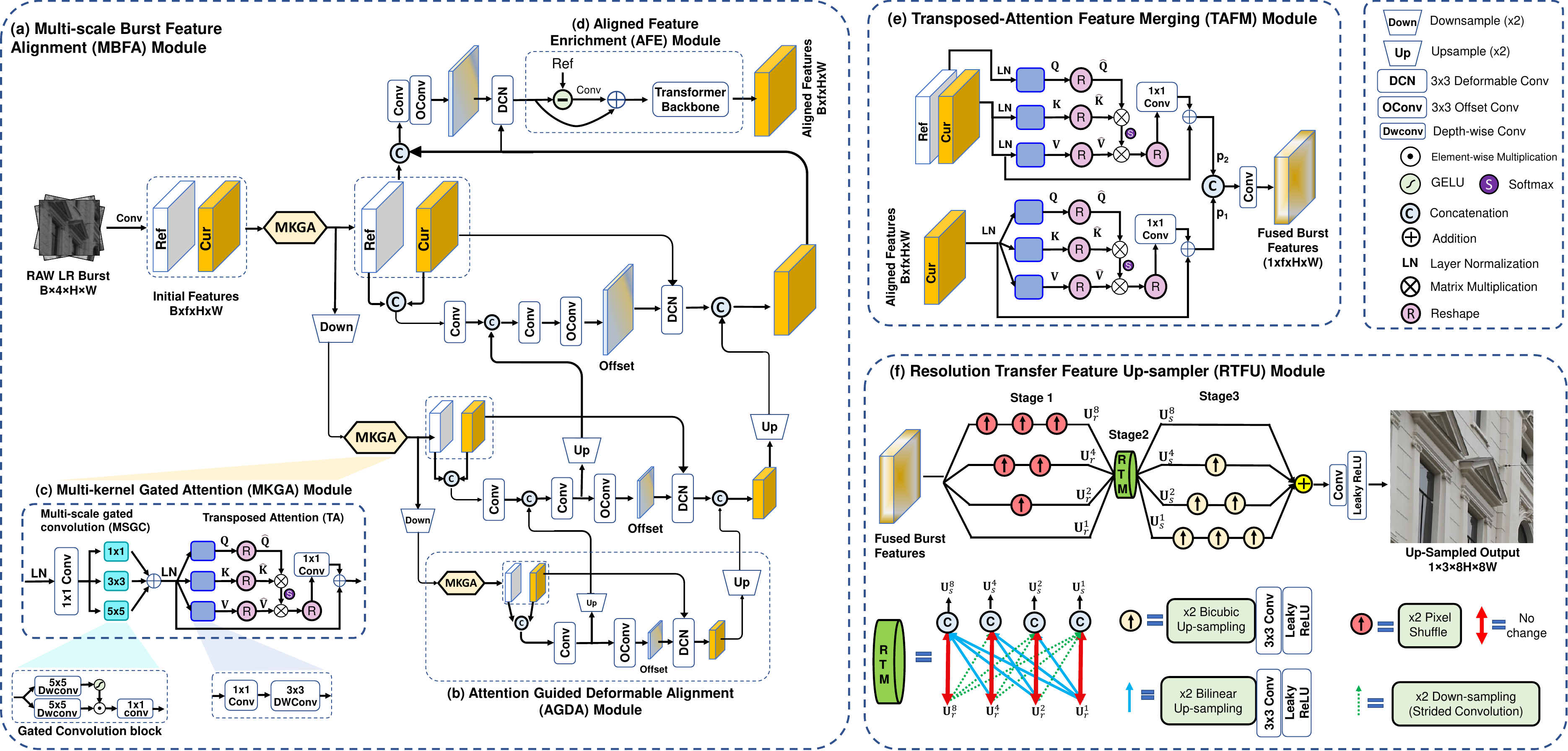}
            \caption{Comprehensive representation of each stage of our proposed GMTNet: (a) The proposed Multi-Scale Burst Feature Alignment (MBFA) module aligns burst features at multiple scales using the proposed (b) Attention-Guided Deformable Alignment (AGDA). The proposed AGDA reduces noise content through our (c) Multi-Kernel Gated Attention (MKGA) module. While, (d) Aligned Feature Enrichment (AFE) boosts high-frequency content through back-projection mechanism and extracts robust features through transformer backbone. (e) Transposed Attention Feature Merging (TAFM) module aggregates the local-non-local pixel interactions within the aligned and reference frames. Lastly, (f) Resolution Transfer Feature Up-sampler (RTFU) up-scales the merged features through single-stage and progressive up-sampling setting using both the conventional and recent up-samplers.}   
            \label{fig: 2}
        \end{figure*}
       Tico \textit{et al.} \cite{tico} leveraged a block-based paradigm, and blocks within and across the burst images are used for performing denoising.
       \cite{noise1, noise2} extended the defacto method of single image denoising approaches, BM3D \cite{dabov} to videos.
       Liu \textit{et al.} \cite{liu1} demonstrated superior denoising performance by using a novel homography flow alignment technique via consistent pixel compositing operator.
       Godard \textit{et al.} \cite{deep} proposed a novel multi-frame denoising model by using burst capture strategy and recurrent deep convolutional neural network.
       Mildenhall \textit{et al.} \cite{gray} introduced Kernel Prediction Network (KPN) to generate per-pixel kernels, utilizing information from multiple images to merge input frames.
       Bhat \textit{et al.} \cite{bhat1} proposed a deep reparametrization of the maximum a posteriori formulation for multi-frame denoising. 
       Dudhane \textit{et al.} \cite{dudhane} proposed a pseudo-burst feature fusion approach for burst frame denoising. 

       \vspace{0.4mm}
    \noindent\textbf{Low-Light Enhancement.}
        Low-light photography in smartphones is limited on account of the small sensor, lens and limited aperture of camera.
        In \cite{learning}, authors introduced a dataset of raw short and long-exposure low-light images, and proposed a learning based pipeline for mapping the degraded low-lit input frames to well-lit sRGB images. 
        Zamir \textit{et al.} \cite{zamir1} proposed a data-driven method for mapping underexposed RAW images to a well-exposed sRGB image. 
        Jung \textit{et al.} \cite{gan} leveraged a novel cycle adversarial network for generating frames in low lighting conditions.
        Liu \textit{et al.} \cite{synthetic} used synthetic events from multiple frames for guiding the enhancement and restoration of low-light frames.  
        Maharjan \textit{et al.} \cite{maharjan} and Zhao \textit{et al.} \cite{zhao}, respectively leveraged a residual learning-based and recurrent convolution network based framework to process burst photos acquired under extremely low-light conditions. 
        Besides super-resolution and denoising, BIPNet \cite{dudhane} is also adept at performing multi-frame low-light image enhancement.
        %

\section{Methodology}
    We present the overall pipeline of our burst processing approach in Figure~\ref{fig: 1}. Given a raw burst image, the goal of our GMTNet is to reconstruct a clean, high-quality image by exploiting the shifted complementary information from the noisy LR image burst. As shown in Figure~\ref{fig: 1}, the input RAW LR burst features are aligned to the reference frame through our proposed \textbf{M}ulti-scale \textbf{B}urst \textbf{F}eature \textbf{A}lignment (MBFA) module. Further, aligned burst features are aggregated using the \textbf{T}ransposed-\textbf{A}ttention \textbf{F}eature \textbf{M}erging (TAFM) module. Lastly, our \textbf{R}esolution \textbf{T}ransfer \textbf{F}eature \textbf{U}p-sampler (RTFU) up-scales the merged features to reconstruct a high-quality image. 
    \subsection{Multi-scale Burst Feature Alignment}
    \label{TPAM}
        Generating an artifact-free, high-quality image through burst processing is highly reliant upon the alignment of the mismatched burst frames.
        However, proper alignment is quite challenging, specifically in low-light and low-resolution images, where noise excessively contaminates the input burst frames.
        Previous burst restoration and video SR methods \cite{dudhane, edvr, bhat1, bhat2, new1,new2} often seek to alleviate these issues by following alignment on locally extracted features. However, they do not explicitly consider the long-range dependencies which are crucial for restoration tasks.
        Consequently, the generated feature maps have limited receptive field making it difficult to align the burst features in case of complex motions.
        We develop the Multi-scale Burst Feature Alignment (MBFA) module to address the mentioned challenges, streamlining burst feature alignment across various scales and facilitating long-range pixel interactions for improved alignment.
        As seen in Figure~\ref{fig: 2}(a), MBFA works in two phases: first, it aligns burst features at multiple scales with the Attention-Guided Deformable Alignment (AGDA) module; second, it refines aligned features via the Aligned Feature Enrichment (AFE) module.
            \subsubsection{Attention-Guided Deformable Alignment} \label{AGDA}
                As discussed in \cite{noise}, noise disturbs the prediction of dense correspondences among multiple frames which is the key concern of several alignment methods.
                However, we find that a well-designed module can easily tackle noisy raw data. 
                Therefore, in order to reduce the noise content in the initial burst features and eventually ease the alignment process, we propose an {A}ttention-{G}uided {D}eformable {A}lignment (AGDA) module that operates at multiple scales to align the burst features as shown in Figure~\ref{fig: 2}(b).
                The proposed AGDA module is inspired from the deformable alignment proposed in TDAN \cite{tdan} and EDVR \cite{edvr}.
                But, their alignment approaches \cite{tdan,edvr} directly apply deformable convolution on the input features, \textit{making them prone to miss the detailed information in case of noisy RAW burst features.}
                Additionally, they also lack at extracting long-range pixel interactions which are useful in complex motions. 
                Our AGDA block addresses these issues by performing implicit feature denoising using MKGA prior to burst feature alignment, instead of directly applying deformable convolution to incoming features.
                Further, the denoised burst features are aligned through the modulated deformable convolution (DCN) as shown in Figure~\ref{fig: 2}(b). 

                \vspace{0.4em}
    \noindent\textbf{Multi-Kernel Gated Attention.}  \label{MKGA}
                The proposed MKGA block offers dynamic adjustment of its receptive field to learn multi-scale \textit{local context} through our Multi-Scale Gated Convolution (MSGC) sub-module and \textit{non-local context} with the Transposed attention (TA) sub-module as demonstrated in Figure~\ref{fig: 2}(c). This adaptability of transitioning between small (local) and large  receptive fields is useful for dealing with various types of image degradation. 
                Given an input tensor $\textbf{Y}$ $\in {\mathbb{R}^{C \times H \times W }}$, the overall operation of MSGC, outputting $\bf{{{\hat Y}}}$  is formulated as:
                \begin{equation}
                    {\bf{\hat Y}} = W_{1}*( {G_{1}({\textbf{Y}})}) + W_{1}*( {G_{3}(\textbf{Y})}) + W_{1}*({G_{5}(\textbf{Y})})
                    \label{eq: 1} 
                \end{equation}
                Here, $W_{1}$ denotes a convolution filter with size 1$\times$1, and * is a convolution operation.
                $G_{k}(\textbf{Y})$ represents the output of the Gated Convolution block \textit{(See Figure~\ref{fig: 2} (c))}, that is mapped out as the element-wise product of two parallel paths for depth-wise convolution layers with filter size \textit{k} and formulated as ${G_k}({\bf{Y}}) = \lambda (W_k^{dep}) \odot W_k^{dep}$. Here, $W_{k}^{dep}$ denotes a depth-wise convolution layer, $ \lambda $ and $\odot$ represents the GELU non-linearity, and element-wise multiplication.

                \vspace{0.4em}
    \noindent \textbf{Transposed Attention.} The extracted multi-kernel features from the MSGC module are passed through the transposed attention (TA) sub-module \textit{(see Figure~\ref{fig: 2}(c))} for capturing their long-range pixel interactions. 
                From a layer normalized tensor $\bf{\hat Y}$, our TA sub-module first generates query \textbf{(Q)}, key \textbf{(K)}, and value \textbf{(V)} projections by applying 1$\times$1 convolutions followed by 3$\times$3 depth-wise convolutions for encoding the non-local and channel-wise spatial context.
                Thereafter, we reshape (\textbf{Q},\textbf{K},\textbf{V}) into ${\bf{\hat Q}}$, ${\bf{\hat K}}$ and ${\bf{\hat V}}$ projections such that the subsequent dot-product interactions between query and key generate a transposed-attention map of size ${\mathbb{R}^{C \times C}}$\cite{restormer}, instead of the huge regular attention map of size ${\mathbb{R}^{HW \times HW}}$ \cite{attention}.
                And, the overall TA process, outputting $\bf{{\tilde Y}}$ is defined as: 
                \begin{equation}
                    \begin{array}{l}
                    {\bf{\tilde Y}} = LN({\bf{\hat Y}}) + {W_1}*(TA({\bf{\hat Q}},{\bf{\hat K}},{\bf{\hat V}}));\\
                    TA({\bf{\hat Q}},{\bf{\hat K}},{\bf{\hat V}}) = {\bf{\hat V}} \otimes S({\bf{\hat K}} \otimes {\bf{\hat Q}})
                    \end{array}
                    \label{eq: 2} 
                \end{equation} 
                Here, $\bf{\hat Y}$ is the feature map obtained from the MSGC module, $LN$ denotes the layer normalization; $TA$ and $S$ denotes the operation of the TA sub-module and Softmax, respectively, ${\bf{\hat Q}} \in {\mathbb{R}^{HW \times C}}$, ${\bf{\hat K}} \in {\mathbb{R}^{C \times HW}}$, and ${\bf{\hat V}} \in {\mathbb{R}^{HW \times C}}$ matrices are obtained after reshaping the tensors from the original size, ${\mathbb{R}^{C \times H \times W}}$, and $\otimes$ denotes matrix multiplication.
                Altogether, the employed MKGA module at each scale allows each pyramidal level to focus on fine details, generating contextualized features that reduce noise and thus ease the subsequent alignment mechanism.
                
                \vspace{0.4 em}
                \noindent \textbf{Modulated Deformable Convolution.} After extracting the features from the MKGA module, we implicitly align the current frame features, ${\bm{ f^b }}$ with the reference frame features (\textit{we considered the first frame as reference}), ${\bm{f^{{b^r}}}} $ via modulated deformable convolution \cite{deformable, tdan} (learnable offsets for deformable convolution layer are obtained through a 3$\times$3 offset convolution layer) as shown in Figure~\ref{fig: 2}(b).
                To ensure better learning, the predicted offsets and aligned burst features are shared from the lower-scale to upper-scale in a bottom-up fashion to ensure semantically stronger and cleaner aligned features. 
            \subsubsection{Aligned Feature Enrichment} \label{AFE}
            \vspace{-0.2cm}
                To fix the remaining minor alignment and noise issues,  we embed a novel Aligned Feature Enrichment (AFE) module on the obtained aligned features.
                The proposed AFE module differs from conventional high-frequency enhancement methods as it extracts local \& non-local features through a transformer backbone, in addition to a back-projection mechanism. This results in a more effective approach for high-frequency enhancement.
                During the back-projection process, we simply compute the high-frequency residue between the aligned burst features and reference frame as shown in Figure~\ref{fig: 2}(d).
                \noindent Thereafter, the local-non-local pixel interactions are enabled by processing the aligned edge boosted burst features through the existing transformer backbone \cite{restormer}. 
                In a nutshell, besides capturing multi-scale local-global representation among the bursts, the AFE module also bridges the gap between the relevant and irrelevant features of the aligned frames.
    \subsection{Transposed-Attention Feature Merging}
    \label{TAFM}
        In burst processing, temporal relation among the multiple frames plays an indispensable role in feature fusion on account of blurry frames from camera perturbations. 
        Considering the fact, that incoming multiple frames have quite a few similar patterns at the feature level, it is infeasible to directly concatenate or add them as it will naively introduce a large amount of redundancy into the network.
        Existing DBSR \cite{bhat1} proposed an attention-based fusion approach but it is limited in exploiting the complementary (global and local) relations that can hinder the information exchange among multiple frames. 
        Further, the recently proposed BIPNet \cite{dudhane} tries to merge the relevant information by concatenating channel-wise features from all burst feature maps. Though it is effective in extracting complementary information, it is computationally extensive.
        \par Unlike the aforementioned fusion techniques, we propose a Transposed-Attention Feature Merging (TAFM) to efficiently encode \textit{inter-frame and intra-frame correlations before merging the frames.}
        As shown in Figure~\ref{fig: 2}(e), TAFM takes queries \textbf{(Q)} and a set of key-value (\textbf{K},\textbf{V}) pairs as input and outputs the linear combination of values that are determined by correlations between the queries and corresponding keys \cite{rnan}.
        The proposed TAFM module has been designed with two parallel blocks (\textit{see Figure~\ref{fig: 2}(e)}), where the lower block (outputting $p_1$) performs the query-key interactions across channels of the aligned neighboring frames to encode the channel-wise local context.
        While the upper block (outputting $p_2$) enhances the feature representations of the reference and current frames by bridging their global correlations. This design allows TAFM to effectively reduce feature redundancy and extract complementary information from multiple frames.
        After encoding the feature correlations globally and locally for a given aligned frame, ${\bm{ {\bar f^b} }}$ with $b$ number of burst frames, the overall merged features of TAFM, ${\bm{ F_m}} \in {\mathbb{R}^{1 \times C \times H \times W}}$ is obtained as follows:
        \begin{equation}
           {\bm{F_m}} = W_{3}*(p_1 \langle C \rangle p_2)
        \end{equation}
        where, $W_{3}$ is a convolution layer with filter size $3\times3$, and $\langle C \rangle$ refers to the concatenation.
    \subsection{Resolution Transfer Feature Up-sampler}
    \label{RTFU}
        The popular up-sampling techniques deployed in SoTA burst SR methods DBSR \cite{bhat1}, DRSR \cite{bhat2} perform direct one-stage up-sampling without leveraging the benefits of information exchange between the HR features and their corresponding LR counterparts. 
        Considering the fact that HR features contain abundant global information and LR features are rich in edge information \cite{zhang,new4,new5}, we design a Resolution Transfer Feature Up-sampler (RTFU) module that is the first upsampler to extract \textit{unique features} of \textit{different resolution spaces.}
        The proposed RTFU module stems from the observation that the transfer of LR and HR features through a multi-resolution framework can be propitious in adaptively recovering the textural information from the fused frames as shown in the ablation study.  
        In RTFU, we target at exploiting the dual benefits of both direct \cite{dong} and progressive up-sampling \cite{lai} strategies using the conventional \cite{bicubic} and recent learnable up-sampling layers \cite{shi} to adequately get into the HR space.
        As shown in Figure~\ref{fig: 2}(f), RTFU achieves its desired HR feature space via a three-stage design: two sets of four parallel progressive multi-resolution streams (Stage1 and Stage3) and a Resolution-Transfer Merging (RTM) module (Stage2).
        \par We first apply progressive up-sampling strategy with pixel-shuffle \cite{shi} (\textit{extreme left of Figure~\ref{fig: 2}(f)}) parallelly in Stage1 for generating ($\times$1, $\times$2, $\times$4, and $\times$8) multi-resolution SR feature responses, which are then forwarded to the RTM module (Stage2). 
        RTM module consists of four input representations: {U$_{r}^i$ (output of Stage1),  {\textit{i} = 1, 2, 4, and 8}} with \textit{i} being the input resolution index, and the associated output representations are given by  U$_{s}^o$, {\textit{o} = 1, 2, 4, and 8} with \textit{o} being the output resolution index.
        Each output representation (U$_{s}^o$) is the concatenation of the transformed representations of the corresponding four inputs \textit{(as shown in the middle of Figure~\ref{fig: 2}(f))}. 
        Thus, the overall operation of Stage2 (RTM module) can be formulated as follows:\newline
        \begin{equation}
            \begin{array}{l}
                U_s^o = {\left[\kern-0.15em\left[ {f(U_r^i)} 
                \right]\kern-0.15em\right]_{i = 1,2,4,8}};\;
                f = \left\{ {\begin{array}{*{20}{c}}
                {1 \,\,\,\,\forall\,\,\,\, i = o}\\
                {\frac{o}{i} \uparrow \,\,\,\,\forall \,\,\,\, o > i}\\
                {\frac{i}{o} \downarrow \,\,\,\,\forall  \,\,\,\, o < i}
                    \end{array}}
                    \right.
            \end{array} 
        \label{eq:5}
        \end{equation}
        Here, $o \in \{ 1,2,4,8\}$, and the mathematical definition of the symbol used in Eq. \ref{eq:5} is given as ${\left[\kern-0.15em\left[ {{A^j}} 
        \right]\kern-0.15em\right]_{j = 1,2,...n}} = {A^1} \langle C \rangle {A^2}.....\langle C \rangle {A^n}$, where $\langle C \rangle$ denotes the concatenation operation among the inputs, and  \textit{f} represents the corresponding transformation operation (upsample or downsample) applied to the input feature $U_{r}$ and is dependent upon the input resolution index (\textit{i}), and the output resolution index (\textit{o}).
        For instance, as shown above, if \textit{o} $>$ \textit{i}, then the corresponding input representation U$_{r}^i$ is up-sampled ($\uparrow$) by a factor of \textit{o}/\textit{i}. In Stage2 (RTM), we deploy bilinear interpolation and strided convolution for feature up-sampling and down-sampling, respectively.
        Thereafter, the resulting features from each branch of Stage3 are again up-sampled progressively using bicubic interpolation to generate an up-sampled feature map of the size ${\mathbb{R}^{1 \times C \times 8H \times 8W}}$. 
        Finally, we add the individual branch output of Stage3 to generate the final high-quality image. 
        Thus, for each pixel location, RTFU can leverage the underlying content information from input frames at multiple-scales and utilize it to get better performance than the mainstream up-sampling operations, pixel-shuffle or interpolations.
\section{Experimental Analysis}
    We validate the proposed GMTNet on real and synthetic datasets for \textbf{(a)} Burst Super-resolution, \textbf{(b)} Burst denoising, and \textbf{(c)} Burst low-light image enhancement tasks.
    \noindent\textbf{Implementation Details.}
        We train separate models for all the considered tasks in an end-to-end manner.
        For better parameter efficiency, we shared each GMTNet module for all burst frames. 
        Our GMTNet has 12.7M parameters with 157 GFLOPs for the burst of size 14$\times$4$\times$48$\times$48 with a running time of 24 fps. To train GMTNet with 4 V100 GPUs, it takes 29 hours for real SR, 97 hours for synthetic SR, 72 hours for grayscale/color denoising, and 38 hrs for burst enhancement.
        All the models are trained with Adam optimizer with \textit{$L_1$} loss function. We employ cosine annealing strategy \cite{sgdr} to decrease the learning rate from 10$^{-4}$ to 10$^{-6}$ during training. 
        For real-world SR, we fine-tune our GMTNet (\textit{with pre-trained weights on SyntheticBurst dataset}) using aligned \textit{L$_1$} loss \cite{bhat1}.
        We provide the task-specific experimental details in the corresponding sections.
        \textit{Additional experimental details and visual results are provided in the supplementary material.}

    \subsection{Burst Super-Resolution}
        We evaluate our proposed GMTNet on synthetic \cite{bhat1, bhat2021ntire} and real-world datasets \cite{bhat1, bhat2021ntire} for scale factor $\times$4.
        Following the settings in \cite{bhat1}, we utilized \textbf{SyntheticBurst} dataset (46,839 and 300 RAW burst sequences for training and validation respectively, where each burst sequence consists of 14 images), and \textbf{BurstSR} dataset consisting of 200 RAW burst sequences (5,405 and 882 patches of size 80$\times$80 for training and validation, respectively).
        
        \begin{table}[t]
            \centering
        \setlength{\tabcolsep}{4pt}
            \caption{\textbf{Burst super-resolution} results for  $\times$4 factor.}
            \scalebox{0.76}{
            \begin{tabular}{l|ccll|ll}
                \toprule
                \multirow{2}{2em}{\textbf{Methods}} & \multicolumn{4}{c|}{\textbf{SyntheticBurst \cite{bhat1}}} & \multicolumn{2}{c}{\textbf{BurstSR \cite{bhat1}}} \\
                 & GFlops & Time (s) & PSNR$\uparrow$ & SSIM$\uparrow$ & PSNR$\uparrow$ & SSIM$\uparrow$ \\
             
                \midrule
                SingleImage & 20.41 & 0.04 & 36.86 & 0.919 & 46.60  & 0.979 \\
                
                WMKPN \cite{weighted} & - & - & 36.56 & 0.912 & 41.87 & 0.958 \\
                HighResNet \cite{deudon} & 400 & 0.05 & 37.45 & 0.924 & 46.64 & 0.980  \\
                DBSR \cite{deep}  & 118 & 0.43 & 40.76 & 0.959 & 48.05 & 0.984  \\
                
                MFIR \cite{bhat2}  & 110 & 0.42 & 41.56 & \textbf{0.964} & 48.33 & 0.985 \\
                
                BIPNet \cite{dudhane} & 300 & 0.13 & \underline{41.93} & {0.960}  & \underline{48.49} & \underline{0.985} \\
                \midrule
                \textbf{Ours}   & 157 & 0.20 & \textbf{42.36} & \underline{0.961} & \textbf{48.95} & \textbf{0.986} \\
                \bottomrule
            \end{tabular}%
            }
            \label{tab:1}%
        \end{table}
        \begin{figure}[t]
            \centering
            \includegraphics[width=1\linewidth]{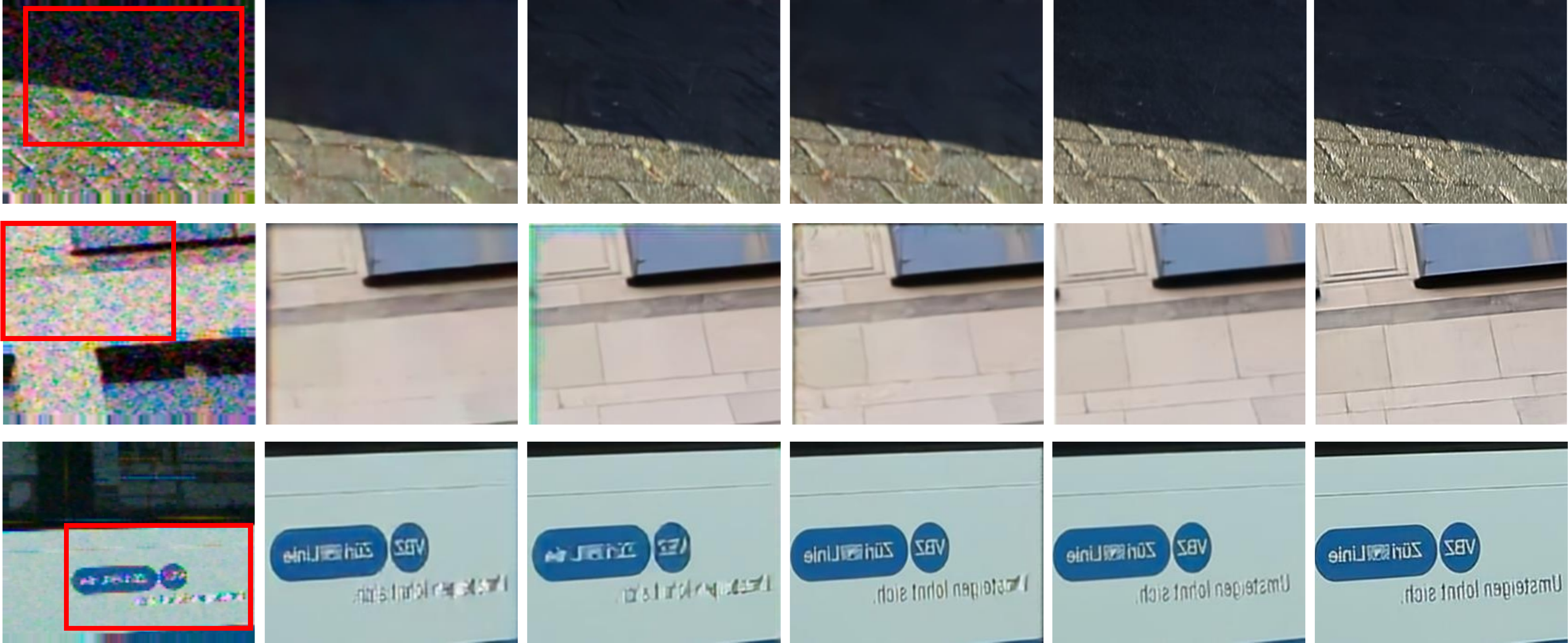}
            \scriptsize Base frame \quad DBSR~\cite{bhat1} \quad LKR~\cite{lecouat2021lucas} \quad \quad MFIR \cite{bhat2} \quad \quad BIPNet \cite{dudhane} \quad {Ours} \quad \quad
            \vspace{1mm}
            \caption{Visual results  on SyntheticBurst~\cite{bhat1} for $\times4$ burst SR.}
            \label{fig: NTIRE_21_syn_val}
        \end{figure}
        \begin{figure}[t]
            \centering
            \includegraphics[width=0.95\linewidth]{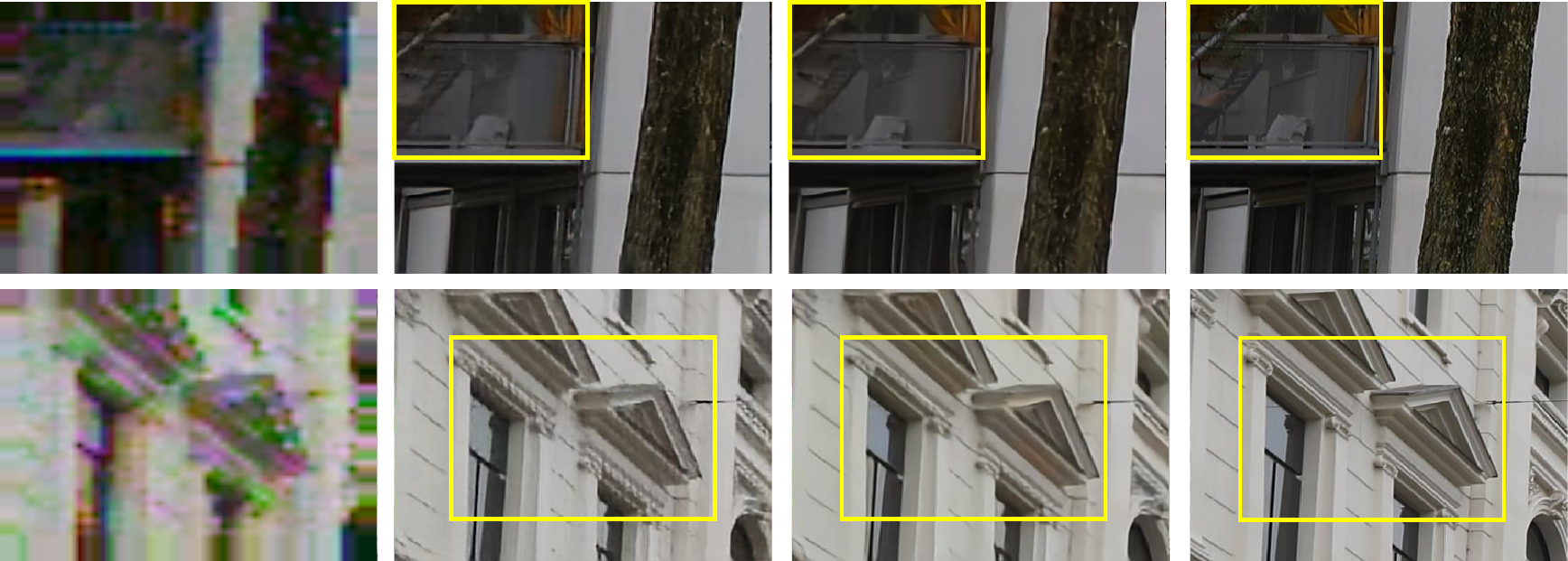}
            \scriptsize Base frame \qquad \qquad BIPNet \cite{dudhane} \qquad \qquad  {Ours} \qquad \qquad Ground Truth
            \vspace{1mm}
            \caption{Visual results on SyntheticBurst~\cite{bhat1} for $\times8$ burst SR.}
            \label{fig: x8_visual_results}
        \end{figure}
        
        \noindent \textbf{SR results on SyntheticBurst dataset for $\times$4  and $\times$8.}
            The proposed GMTNet is trained for 300 epochs on the training split of SyntheticBurst dataset for both $\times4$, and $\times8$ up-sampling tasks and evaluated on the validation set of SyntheticBurst dataset \cite{bhat1}.
            We compared our proposed GMTNet with several SoTA approaches for $\times$4 as shown in Table \ref{tab:1}.
            Particularly, our GMTNet obtains a PSNR gain of about 0.43 dB over the previously best-performing BIPNet \cite{dudhane} and 0.80 dB over the second-best approach \cite{bhat2}.
            To further prove the potency of our proposed GMTNet on large scale factors, we conduct an experiment for $\times$8 burst SR.
            The LR-HR pairs are synthetically generated using the same procedure described for SyntheticBurst dataset \cite{bhat1}. Visual results shown for a few challenging images in Figure~\ref{fig: NTIRE_21_syn_val} ($\times$4) and Figure~\ref{fig: x8_visual_results} ($\times$8) clearly prove that results obtained by GMTNet are sharper and it efficiently reconstructs the structural content and fine textures, without compromising details. In Table \ref{tab:1}, we also compare the computational complexity  of several state-of-the-art burst SR methods. 
        \begin{figure}[t]
            \centering
            \includegraphics[width=1\linewidth]{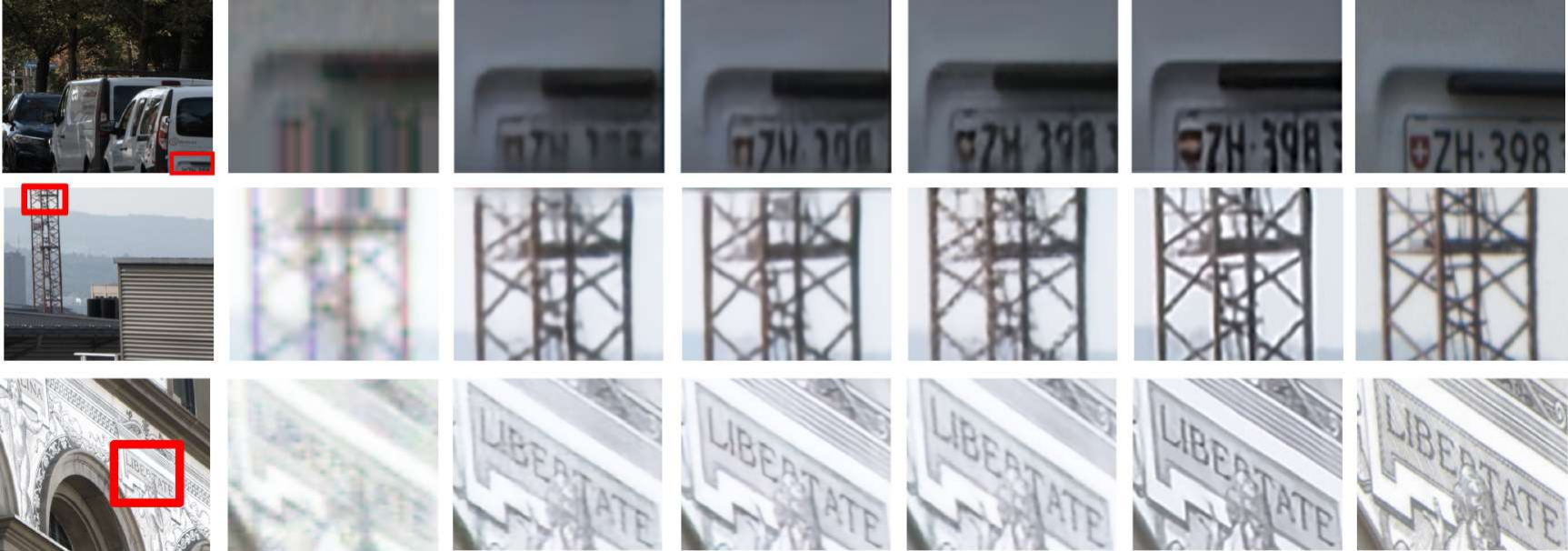}
            \scriptsize HR Image \, Base frame \, DBSR~\cite{bhat1} \,\,  MFIR~\cite{bhat2} \, BIPNet \cite{dudhane} \,\,{Ours} \,\,\,\,\,\,\,\ GT-Image
            \vspace{1mm}
            \caption{Results on real BurstSR dataset~\cite{bhat1} for $\times4$ burst SR.}
            \label{fig: NTIRE_21_real_val}
        \end{figure}
        
        \noindent \textbf{SR results on BurstSR dataset.} Since, the LR-HR pairs for BurstSR dataset are captured using different cameras, they suffer from minor misalignment. Thus we follow the previous work \cite{bhat1} and use aligned \textit{L$_1$} loss for fine-tuning the GMTNet for 25 epochs and evaluate our model by using aligned PSNR/SSIM. Table \ref{tab:1} shows that our proposed GMTNet obtain conducive results, outperforming SoTA BIPNet \cite{dudhane} by a substantial gain of 0.46 dB. Visual comparisons in Figure~\ref{fig: NTIRE_21_real_val} depict that unlike other compared methods, the proposed GMTNet is more effective for generating minute details in the reconstructed images, with better color and structure preservation. 
        \begin{figure}[t]
            \centering
            \includegraphics[width=1\linewidth]{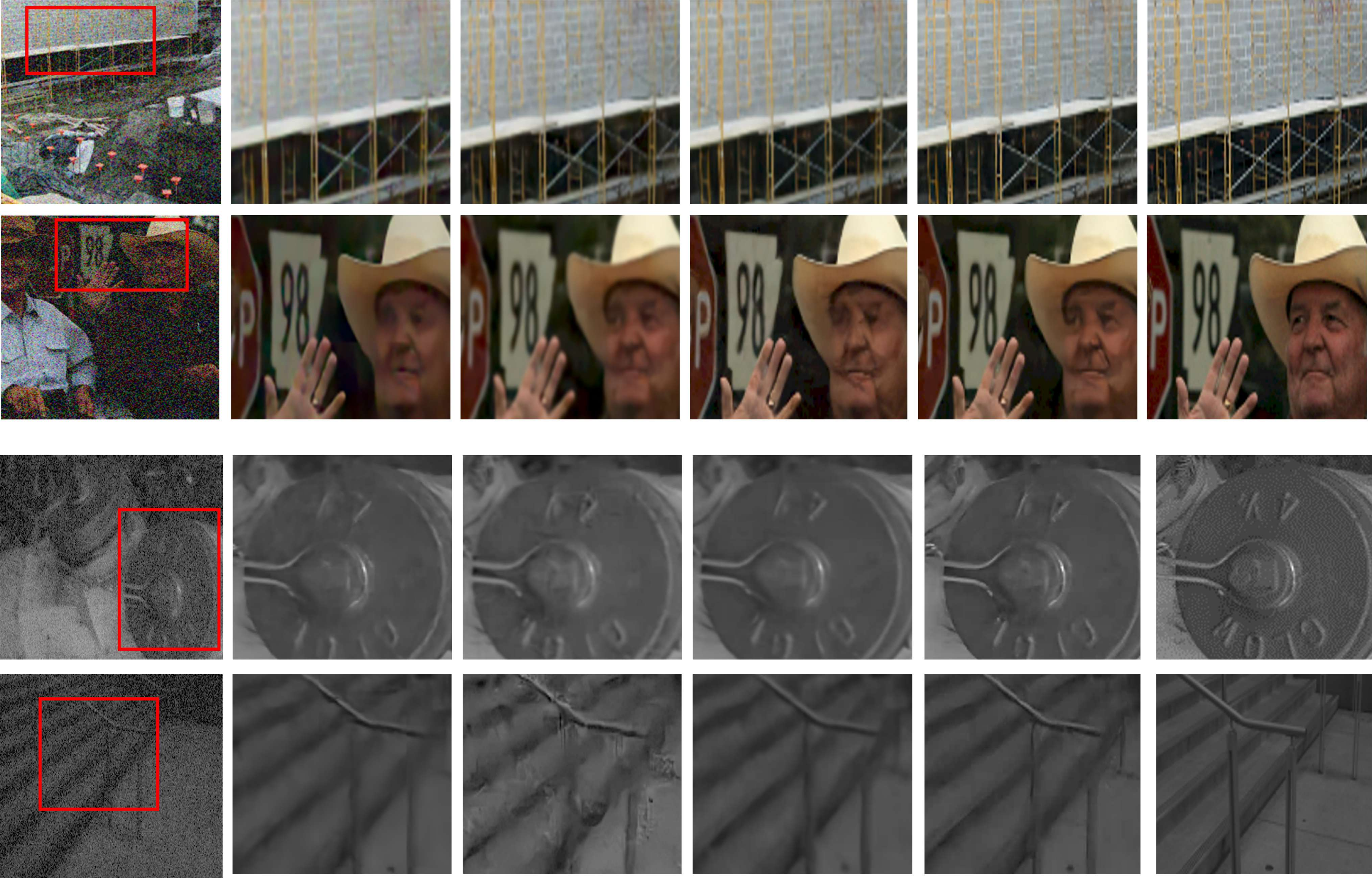}
            \scriptsize HR Image \quad  \quad BPN \cite{color} \quad \quad MFIR \cite{bhat2} \quad BIPNet \cite{dudhane} \quad  \quad {Ours}  \quad \quad GT-Image 
            \vspace{1mm}
            \caption{ Visual results on color datasets~\cite{color} (first two rows) and gray-scale~\cite{gray} (last two rows) for burst denoising.}
            \label{fig: burst_denoising}
        \end{figure}

    \subsection{Burst Denoising Results}
        This section presents the results of burst denoising on color (test split: 100 bursts) \cite{color} as well as gray-scale (test split: 73 bursts) \cite{gray} datasets. Both these datasets have four variants with different noise gains $(\text{1, 2, 4, 8})$, corresponding to noise parameters $(\log(\sigma_r), \log(\sigma_s)) \rightarrow$ $ (\text{-2.2, -2.6})$, $(\text{-1.8, -2.2})$, $(\text{-1.4, -1.8})$, and $(\text{-1.1, -1.5})$, respectively. We train grayscale and color burst denoising models for 200 epochs on 20k synthetic noisy samples (generated as in \cite{bhat2}).
        \begin{table}[htbp]
        	\centering
            \caption{\textbf{Gray-scale burst denoising}~\cite{gray} results with PSNR.}
        	\centering
                \setlength{\tabcolsep}{2pt}
                \scalebox{0.9}{
        		\begin{tabular}{lccccc}
                    \toprule
                    \textbf{Methods} & Gain $\propto$ 1 & Gain $\propto$ 2 & Gain $\propto$ 4 & Gain $\propto$ 8 & Average\\
                    \midrule
                    KPN~\cite{gray}&36.47&33.93&31.19&27.97&32.19\\
                    BPN~\cite{color}& 38.18 & 35.42 &32.54& 29.45 & 33.90\\
                    BIPNet~\cite{dudhane} & 38.53 & 35.94 & 33.08 & 29.89 & 34.36 \\
                    MFIR~\cite{bhat1} & \textbf{39.37} & \textbf{36.51} & 33.38 & 29.69 & 34.74 \\
                    \midrule
                    \textbf{Ours} & 39.07 & 36.46 & \textbf{33.52} & \textbf{30.46} & \textbf{34.87} \\
                    \bottomrule
                \end{tabular}
        	}
           	\label{tab:kpn_grayscale}%
        	
        \end{table}

        \begin{table}[t]
        	\centering
        	\caption{\textbf{Color burst denoising} ~\cite{color} results with PSNR.}
        	\setlength{\tabcolsep}{2pt}
                \scalebox{0.9}{
        		\begin{tabular}{lccccc}
                    \toprule
                    \textbf{Methods}&Gain $\propto$ 1 & Gain $\propto$ 2 & Gain $\propto$ 4 & Gain $\propto$ 8 & Average \\
                    \midrule
                    KPN~\cite{gray} & 38.86 & 35.97 & 32.79 & 30.01 & 34.40 \\
                    BPN~\cite{color} & 40.16 & 37.08 & 33.81 & 31.19 & 35.56 \\
                    BIPNet~\cite{dudhane} & 40.58 & 38.13 & 35.30 & 32.87 & 36.72\\
                    MFIR~\cite{bhat1} & \textbf{41.90} & 38.85 & 35.48 & 32.29 & 37.13 \\
                    \midrule
                    \textbf{Ours} & 41.74 & \textbf{38.91} & \textbf{35.74} & \textbf{33.09} & \textbf{37.38} \\
                    \bottomrule
                \end{tabular}
        	}
        	\label{tab:kpn_color}%
        \end{table}
        \noindent \newline \textbf{Denoising results.}
            Table \ref{tab:kpn_grayscale} shows the results on the gray-scale burst denoising dataset against SoTA methods. Our GMTNet outperforms the recent BIPNet\footnote{Existing BIPNet results are collected from their official GitHub repository.} \cite{dudhane} by about 0.57 dB for the highest noise gain (Gain $\propto$ 8). Similarly, for color denoising, our approach outperforms existing MFIR \cite{bhat1} on all four noise levels (except the lowest noise gain) with an average margin of 0.25 dB as shown in Table \ref{tab:kpn_color}. Qualitative comparison in Figure~\ref{fig: burst_denoising} clearly proves the efficacy of our approach in recovering the required subtle contextual details, thus generating cleaner denoised outputs. 
        
        \begin{figure}[t]
            \centering
            \includegraphics[width=1\linewidth]{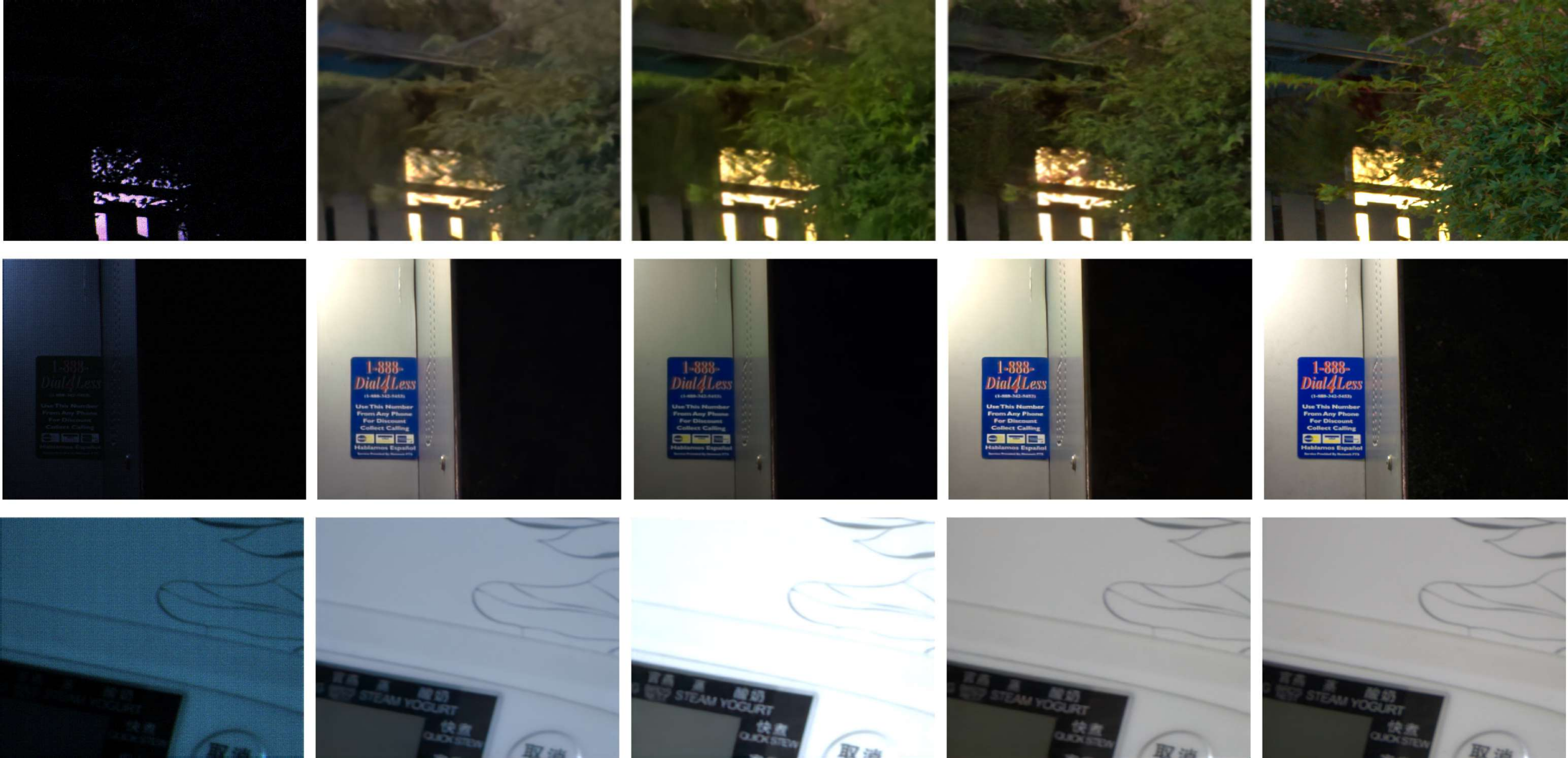}
            \scriptsize RAW input  \quad \quad    LLED \cite{karadeniz} \quad \quad  BIPNet \cite{dudhane}  \qquad Ours \qquad \qquad {GT-image} 
            \vspace{1mm}
            \caption{Visual results on SONY-subset of SID dataset~\cite{learning} for burst low-light image enhancement.}
            \label{fig:enhancement}
        \end{figure}
    \subsection{Low-Light Enhancement Results}
        
        \begin{table}
            \centering
            \setlength{\tabcolsep}{8pt}
            \caption{\textbf{Burst low-light enhancement} on Sony-subset~\cite{learning}.}
            \scalebox{0.9}{
            \begin{tabular}{lccc}
                \toprule
                \textbf{Methods} & \textbf{PSNR $\uparrow$} & \textbf{SSIM $\uparrow$} & \textbf{LPIPS $\downarrow$} \\
                \midrule
                Chen \textit{et al.}\cite{learning} & 29.38 & 0.89 & 0.48 \\
                Maharjan \textit{et al.} \cite{maharjan} & 29.57 & 0.89 & 0.48 \\
                Zamir \textit{et al.} \cite{zamir1} & 29.13 & 0.88 & 0.46 \\
                Zhao \textit{et al.} \cite{zhao} & 29.49 & 0.89 & 0.45 \\
                Karadeniz \textit{et al.} \cite{karadeniz} & 29.80 & 0.89 & 0.30 \\
                BIPNet~\cite{dudhane} & 32.87 & 0.93 & \textbf{0.30} \\
                \midrule
                \textbf{Ours} & \textbf{33.13} & \textbf{0.94} & {0.31} \\
                \bottomrule
            \end{tabular}
            }
            \label{tab: enhancement}
        \end{table}
        Following other existing works \cite{dudhane,karadeniz}, we test the performance of our GMTNet on the SONY-subset from the SID dataset \cite{learning}.
        It contains 161 input RAW burst sequences for training, 36 for validation, and 93 for testing.  
        We train the proposed GMTNet with $L_1$ loss for 200 epochs on 5000 cropped patches of size 256$\times$256 from the training set of SONY-subset.
        Table~\ref{tab: enhancement} gives the image quality scores for several competing approaches.
        The proposed GMTNet provides 0.26 dB improvement over the existing best BIPNet \cite{dudhane}.  
        Visual comparisons in Figure~\ref{fig:enhancement} show that the enhanced images are relatively cleaner, sharper and preserves more structural content than other compared approaches.
        
        \begin{table}[t]
            \centering
            \footnotesize
            \caption{Ablation study for GMTNet contributions. PSNR is reported on SyntheticBurst dataset~\cite{bhat1} for $\times$4 burst SR task.}
            \setlength{\tabcolsep}{4pt}
            \scalebox{0.95}{
            \begin{tabular}{l|l@{$\;\,$}c@{$\;\,$}c@{$\;\,$}c@{$\;\,$}c@{$\;\,$}c@{$\;\,$}c@{$\;\,$}c@{$\;\,$}c@{$\;\,$}}
                \toprule
                \textbf{Task} & \textbf{Modules} & &  &  &  &  & & &\\
                \toprule
                & Baseline & \checkmark     & \checkmark     & \checkmark     & \checkmark     & \checkmark     & \checkmark     & \checkmark  & \checkmark   \\
                \midrule
                \multirow{3}{2.2em}{\scriptsize \textbf{Align ment} (\textsection\ref{TPAM})} &  w/O MKGA &       & \checkmark     &   &      &      &     &  &    \\
                & with MKGA &         &     & \checkmark     & \checkmark     & \checkmark     & \checkmark     & \checkmark & \checkmark \\
                & AFE &       &         &     & \checkmark     & \checkmark     & \checkmark     & \checkmark & \checkmark \\
                \midrule
                \multirow{3}{2.2em}{\scriptsize \textbf{Fusion} (\textsection\ref{TAFM}) } & with p$_1$ &           &       &       &      & \checkmark     &      &  & \\
                & with p$_2$ &       &       &       &    &  & \checkmark     &   &  \\
                & with p$_1$+p$_2$ &       &       &       &    &  &     & \checkmark  & \checkmark \\
                \midrule
                \multicolumn{2}{l}{\textbf{Upsample} (\textsection\ref{RTFU})}  &       &       &       &       &       &     &   & \checkmark \\
                \midrule
                & \textbf{PSNR} & 36.38 & 38.02 & 39.12 & 39.40 & 39.84 & 40.23 & 40.74 & \textbf{41.82}\\
                \bottomrule
            \end{tabular}}
            \label{tab: ablations1}%
            
        \end{table}

\section{Ablation Study}
    Here we analyze the influence of every key component and design choice in our formulation. All models are trained for 100 epochs on SyntheticBurst dataset~\cite{bhat1} for $\times$4 burst SR task.
    As reported in Table \ref{tab: ablations1}, the baseline model achieves a PSNR of 36.38 dB. For the baseline model we deploy addition operation for fusion and pixel-shuffle for up-sampling.
    After adding the proposed modules to the baseline network, the results improve persistently and notably.
    For instance, we attain a performance gain of 3.02 dB when we incorporate our alignment module into the baseline model.
    The insertion of the proposed fusion and up-sampling modules in our network further improves the PSNR of the overall network by about 1.34 dB and 1.08 dB, respectively.
    Overall, GMTNet obtains a compelling gain of 5.44 dB over the baseline model. 

\begin{figure}[t]
        \centering
        \includegraphics[width=0.95\linewidth]{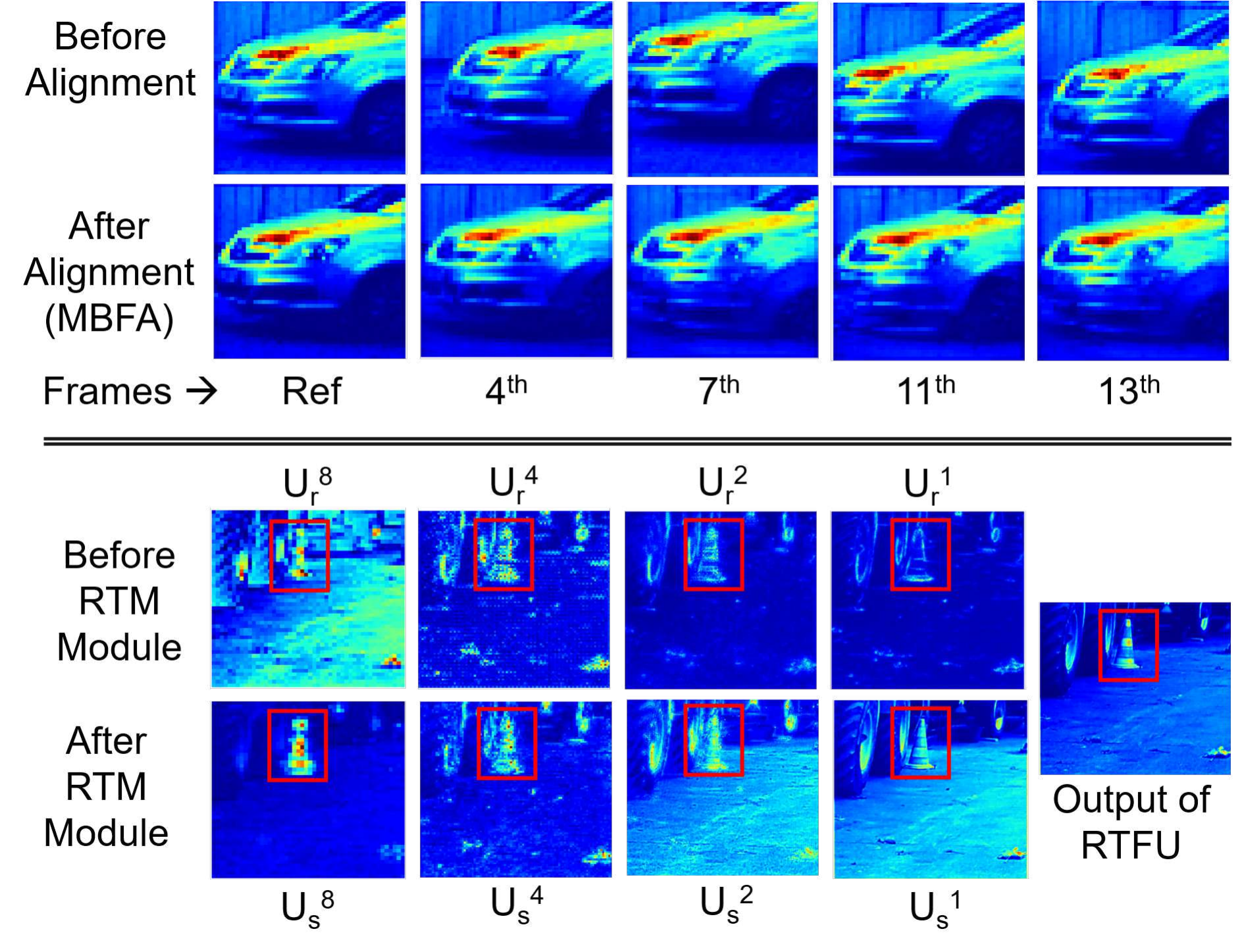}
        \caption{Feature map visualizations before and after applying proposed MBFA (Figure \ref{fig: 2}(b)) and RTM (middle of Figure \ref{fig: 2}(f)) modules into our GMTNet.} 
        \label{fig: feat_alignment}
    \end{figure}

    \noindent\textbf{Effectiveness of MBFA module.}
        As reported in Table \ref{tab: ablations1}, the inclusion of MKGA and AFE modules into our alignment (MBFA) module provides a performance boost of around 1.10 dB and 0.28 dB, respectively which supports the effectiveness of the proposed modules in capturing motion cues. Further, we compare the GMTNet results in Table \ref{tab:ablation2} (a) by replacing MBFA with other popular explicit and implicit alignment approaches \textit{(Keeping the rest of the modules same)}. We observe that the MBFA module obtains a performance gain of about 0.83 dB over PCD module proposed in EDVR~\cite{edvr}. To further highlight the ability of MBFA module in aligning burst features, we visualize the features (of few frames) before and after applying it as shown in Figure~\ref{fig: feat_alignment}. It clearly reveals our MBFA works well without any dedicated supervision.
        \begin{table}[t]
            \centering
            \caption{Impact of the proposed modules in terms of PSNR/SSIM on SyntheticBurst SR dataset for $\times$4 burst SR task. }
            \setlength{\tabcolsep}{4pt}
            \scalebox{0.9}{
            \begin{tabular}{clll}
                \toprule
                \textbf{Task} & \textbf{Methods} & \textbf{PSNR$\uparrow$} & \textbf{SSIM$\uparrow$}\\
                
                \midrule
            
                \multirow{4}{7em}{\textbf{(a) Alignment}}
                    & GMTNet + PCD \cite{edvr}  & 40.99 & 0.953 \\
                    & GMTNet + Explicit \cite{bhat1} & 39.26 & 0.944 \\
                    & GMTNet + EBFA \cite{dudhane} & 41.10 & 0.958  \\
                    & \textbf{GMTNet} + \textbf{MBFA} & \textbf{41.82} & \textbf{0.960} \\
                
                \midrule
            
                \multirow{4}{7em}{\textbf{(b) Burst Fusion}} 
                    & GMTNet + TSA \cite{edvr} & 39.97 & 0.947  \\
                    & GMTNet + DBSR \cite{deep} & 40.32 & 0.950 \\
                    & GMTNet + PBFF \cite{dudhane} & 41.60 & 0.954  \\
                    & \textbf{GMTNet} + \textbf{TAFM} & \textbf{41.82} & \textbf{0.960} \\
            
                \midrule
            
                \multirow{4}{7em}{\textbf{(c) Upsampler}}
                    & GMTNet + Bil & 40.22 & 0.940 \\
                    & GMTNet + PS \cite{deep} & 40.41 & 0.943  \\
                    & GMTNet + AGU \cite{dudhane} & 41.30 & 0.951  \\
                    & \textbf{GMTNet} + \textbf{RTFU} & \textbf{41.82} & \textbf{0.960} \\
                \bottomrule
            \end{tabular}%
            }
            \label{tab:ablation2}%
        \end{table}%

    \noindent \textbf{How to design TAFM module?}
        A trivial design of our TAFM module is to use a single stream for extracting the information and then concatenating the features. However, from Table \ref{tab: ablations1}, it is clear that utilizing both the p${_1}$ and p${_2}$ outputs for subsequent merging results in a performance boost of around 0.90 dB. It clearly signifies that two-stream TAFM performs better than any single-stream.
    
    \noindent\textbf{Impact of TAFM module.}
        The results in Table \ref{tab:ablation2} for burst fusion tasks further show that replacing our TAFM module with other popular fusion modules have a detrimental influence on the overall performance of our model, with PSNR drop of around 0.22 dB when utilizing the recently proposed PBFF \cite{dudhane} module in our network.
    \noindent\textbf{Effectiveness of the proposed RTFU.}
        To validate the effectiveness of our RTFU, we replace it with the conventional and recent, bilinear interpolation (Bil) and pixel-shuffle (PS), AGU respectively. The accuracy scores in Table \ref{tab:ablation2}, clearly demonstrate its ability to reconstruct a high-quality image.

    \noindent\textbf{How important is the proposed RTM module in RTFU?}
        To prove the imperativeness of the RTM module, in Figure~\ref{fig: feat_alignment} we visualize the feature maps before and after embedding it in RTFU. It clearly proves that our model benefits from the efficient use of both LR and HR information to complete the restoration of sharp regions. 
\section{Conclusion}
    We present a generalised network for burst processing to reconstruct a single high-quality image from a given RAW burst of low-quality noisy images.
    In the proposed approach, our Multi-scale Burst Feature Alignment (MBFA) module aligns the noisy burst features at multiple scales using the proposed Attention-Guided Deformable Alignment (AGDA).
    The inclusion of Aligned Feature Enrichment (AFE) module improves the aligned features by fixing any minor misalignment issue, thus yielding well-refined, denoised and aligned features.
    To further improve model robustness, Transposed Attention Feature Merging (TAFM) module manifests efficient fusion performance by analyzing the global and local correlations among the incoming frames. 
    Finally, the proposed Resolution Transfer Feature Up-sampler (RTFU) up-scales the merged features by consolidating information from both LR and HR feature spaces to reconstruct a high-quality image.
    Consistent achievement of SoTA results for burst super-resolution, denoising and low-light enhancement on synthetic and real datasets corroborates the robustness and potency of our approach.

{\small
\bibliographystyle{ieee_fullname}
\bibliography{arxive}
}

\renewcommand\thesection{\Alph{section}}
\setcounter{section}{0}

\setcounter{table}{0}
\renewcommand{\thetable}{S\arabic{table}}

\setcounter{figure}{0}
\renewcommand{\thefigure}{S\arabic{figure}}

\onecolumn
\begin{center}
\textbf{\LARGE Supplemental Material}
\end{center}
\vspace{2cm}

This supplementary material contains:
 \begin{itemize}
     \item Detailed explanation of the datasets (\textsection \ref{dataset}).
     \item Additional Ablation Study (\textsection \ref{ablation}).
     \item Difference with prior works (\textsection \ref{difference}).
     \item More qualitative results (\textsection \ref{results}).
     \item Feature Map Visualizations (\textsection \ref{visualization}).
     \item Future Work (\textsection \ref{future}).
 \end{itemize}
 
\section{Dataset Details} 
\label{dataset}
\subsection{Burst Super-resolution}
(1) \textbf{SyntheticBurst dataset} consists of 46,839 and 300 RAW bursts for training and validation, respectively. %
Each burst contains 14 LR RAW images generated synthetically from a single sRGB image (each of size 48$\times$48 pixels). 
Each sRGB image is first converted to RAW camera space using the inverse pipeline \cite{bhat1}. 
Next, the burst is generated with random translations and rotations. 
Finally, the LR burst is obtained by applying the bilinear downsampling followed by Bayer mosiacking, sampling and noise addition operations.

(2) \textbf{ BurstSR} dataset \cite{bhat1} consists of 200 real burst images, with each having 14 LR RAW images. 
LR-HR pair for this dataset has been acquired with  Samsung Galaxy S8 smartphone camera and DSLR camera, respectively. 
From the acquired 200 RAW burst sequences, crops of spatial size 80$\times$80 have been extracted for obtaining a training set consisting of 5405 images and validation dataset comprising of 882 images.

\subsection{Burst Denoising}
Following the experimental settings of \cite{bhat2}, we utilize 20k samples from the Open Images \cite{krasin2017openimages} training set to generate the synthetic noisy bursts of burst size 8 and spatial size of 128$\times$128.
We evaluate our approach on the grayscale and color burst denoising datasets in \cite{gray} and \cite{color}.
Both these datasets contains 73 and 100 bursts, respectively. For both these datasets, a burst is generated synthetically by applying random translations to the base image. 
The shifted images are then corrupted by adding heteroscedastic Gaussian noise with variance $\sigma _r^2 + {\sigma _s}x$. 
Here, $x$ denotes the clean pixel value, and $(\sigma_r, \sigma_s)$ denotes the read and shot noise parameters, respectively.
While training, the noise parameters $(\log(\sigma_r), \log(\sigma_s))$ are sampled uniformly in the log-domain from the range $\log(\sigma_r) \in[-3,-1.5]$ and $\log(\sigma_s) \in[-4,-2]$. 
The proposed GMTNet is then evaluated on 4 different noise gains $(\text{1, 2, 4 and 8})$ corresponding to the noise parameters $(\log(\sigma_r), \log(\sigma_s)) \rightarrow$ $ (\text{-2.2, -2.6})$, $(\text{-1.8, -2.2})$, $(\text{-1.4, -1.8})$, and $(\text{-1.1, -1.5})$, respectively.
\textit{Note that the noise parameters for the highest noise gain, \ie, 8 are unseen during training.}
Therefore, the performance of this noise level is an indication of the generalization of the network to unseen noise.

\subsection{Burst Low-light enhancement}
We train and evaluate our model on the See-in-the-Dark (SID) dataset \cite{learning}, that consists of short exposure burst raw images taken under extremely dark indoor (0.2-5 lux) or outdoor (0.03-0.3 lux) scenes.
All these images are acquired with three different exposure times of 1/10, 1/25 and 1/30 seconds, where the corresponding reference images are obtained with 10 or 30 seconds exposures depending on the scene. 
We evaluate the performance of our models on the Sony subset, that contains 161, 36 and 93 distinct burst sequences for training, validation and testing, respectively. The number of burst images varies from 2-10 for every distinct scene.

\section{Additional Ablation Study} 
\label{ablation}
\subsection{Effect of our resolution transfer and aligned feature enrichment modules}
\begin{figure}[htbp]
            \centering
            \includegraphics[width=1\linewidth]{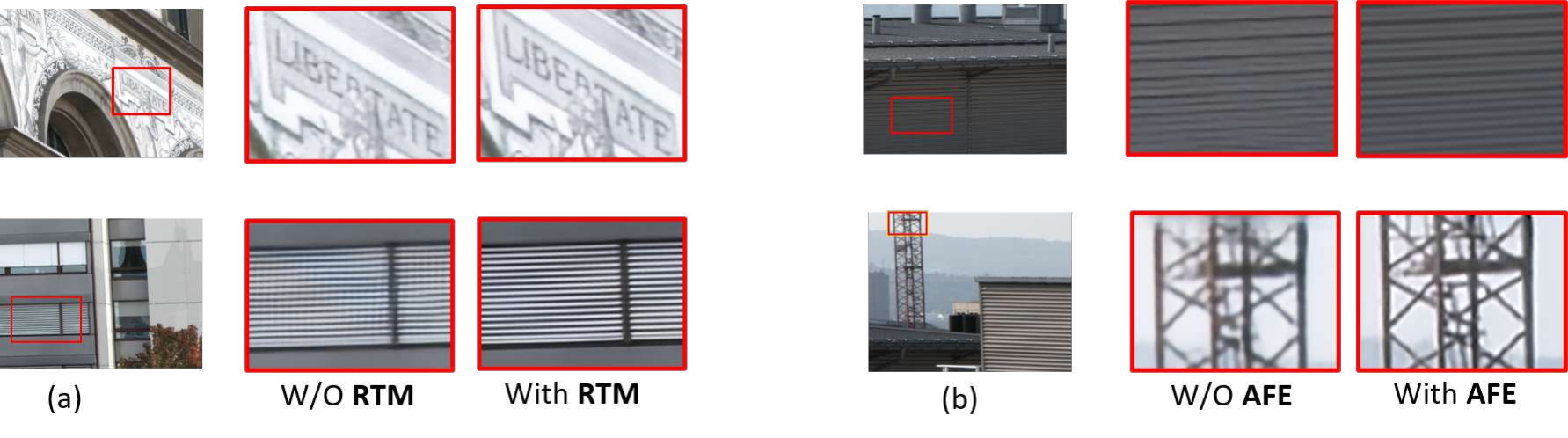}
             \scriptsize
            \vspace{-2mm}
            \caption{(a) Qualitative Comparison on BurstSR dataset without (W/O) and with Resolution Transfer Merging (RTM) module. Our module benefits from the efficient extraction of features in both low and high-resolution space and makes effective use of information to complete the restoration of the sharp regions, (b) Effect of the proposed aligned feature enrichment (AFE) module in our proposed GMTNet. Images recovered by employing AFE module have lesser artifacts when compared to images obtained without it. }
            \label{fig: ablation_1}
        \end{figure}

As shown in Figure \ref{fig: ablation_1}, without the proposed resolution transfer module (RTM) and the aligned feature enrichment (AFE) module, the super-resolution results on real-world photos are blurry. 

\section{Difference with prior works} 
\label{difference}
    In this work, we propose a network that jointly performs denoising, demosaicking and super-resolution. 
    We contribute in all the three components of burst processing \textit{i.e.} alignment, fusion and up-sampling.
    Table \ref{tab:table1} highlights the differences between the popular burst restoration approaches \cite{edvr, bhat1, dudhane} and our GMTNet. 
    First, in contrast to the existing alignment modules, the proposed multi-scale burst feature alignment (MBFA) approach denoises and implicitly aligns the burst features at multiple scales using our multi-kernel gated attention (MKGA) and attention-guided deformable alignment (AGDA) modules, respectively.
    Our MBFA also enriches the aligned burst features through back-projection mechanism and extracts the local and non-local features via encoder-decoder based transformer. 
    As shown in Table \ref{tab:table1}, existing DBSR~\cite{bhat1}, EDVR~\cite{edvr} and BIPNet~\cite{dudhane} lack some of these properties: feature denoising, multi-scale feature alignment, local-non-local feature extraction, back-projection (feature enrichment), and implicit feature alignment.  
    
    \par Keeping in consideration that different neighboring frames are not equally informative due to occlusion/blurry regions, and misalignment arising from the preceding alignment stage may adversely affect the reconstruction performance, we propose the \textit{transposed attention feature merging} module to adaptively pay attention to the current frame, that is more similar to the reference frame. 
    To capture the similarity between frames, EDVR~\cite{edvr} process the incoming frames in a pyramidal manner through a temporal and spatial attention-based fusion module. 
    This approach \cite{edvr} focuses more on correlating the frames locally and does not consider the long-range pixel interactions as well as inter-dependencies among different channels.
    BIPNet~\cite{dudhane} propose the pseudo burst mechanism to adaptively merge the frames by considering the inter-dependencies between the channels, but it is quite computationally extensive. 
    The weighted summation-based fusion approach in DBSR~\cite{bhat1} applies a softmax function onto the aligned frames, and reference features to attentively assign weights to the aligned and misaligned regions. 
    However, none of the above approaches considers the merits of computing both the \textbf{local and non-local correlations} among the incoming frames.
    Our two-stage parallel strategy targets at computing both these relations among the frames by considering \textbf{long-range pixel dependencies} between the reference and supporting frames and also within the supporting frames to ease the fusion process.  
    \par The up-sampling module generates the output high-resolution image from the fused feature map. 
    Unlike the compared approaches that utilize either \textbf{conventional} or \textbf{recent up-sampling techniques}, our proposed resolution transfer feature up-sampler (RTFU) considers the merits of utilizing both recent \textit{(slow and more accurate)} and conventional up-samplers \textit{(fast and less accurate)} to get into high-resolution space.
    Additionally, as shown in the Table \ref{tab:table1}, the up-sampling strategies in burst restoration methods either use \textbf{progressive} up-sampling \textit{(for generating intermediate SR predictions at multiple resolutions)} or \textbf{direct} up-sampling, but unlike ours, none of these approaches combines both these strategies in a single network.
    
    \begin{table*}[htbp]
  \centering
  \scriptsize
  \setlength{\tabcolsep}{4pt}
  \caption{Comparison between the proposed and existing alignment, fusion and up-sampling modules based on their \textbf{properties.}}
    \begin{tabular}{p{1.5cm}|p{2.3cm}p{2.3cm}p{4cm}p{5cm}}
    \toprule
        Tasks  &  \multicolumn{1}{c}{\textbf{DBSR\cite{bhat1}}} &  \multicolumn{1}{c}{\textbf{EDVR\cite{edvr}}} &   \multicolumn{1}{c}{\textbf{BIPNet\cite{dudhane}}} &  \multicolumn{1}{c}{\textbf{Ours}} \\
    \midrule
    Alignment & Explicit multi-scale approach, local feature extraction & Implicit multi-scale approach, local feature extraction & Implicit single-scale approach, denoising,  local feature extraction, back-projection & {Implicit multi-scale approach, denoising, local-non-local feature extraction, back-projection, feature refinement} \\ \midrule
    Fusion & Weighted summation, local feature extraction  & Local feature correlation & Inter-frame interaction, local feature extraction & {Inter-frame interaction, local-non-local feature correlations} \\ \midrule
    Up-sampling & Direct, pixel-shuffle & Direct, bilinear interpolation & Progressive, transposed convolution & {Direct, progressive, pixel-shuffle, bilinear, bicubic interpolation, resolution-transfer} \\ 
    \bottomrule
    \end{tabular}%
  \label{tab:table1}%
\end{table*}%
    \section{Additional Visual Results} 
    \label{results}
\begin{figure*}[t]
    \centering
    \includegraphics[width=1\linewidth]{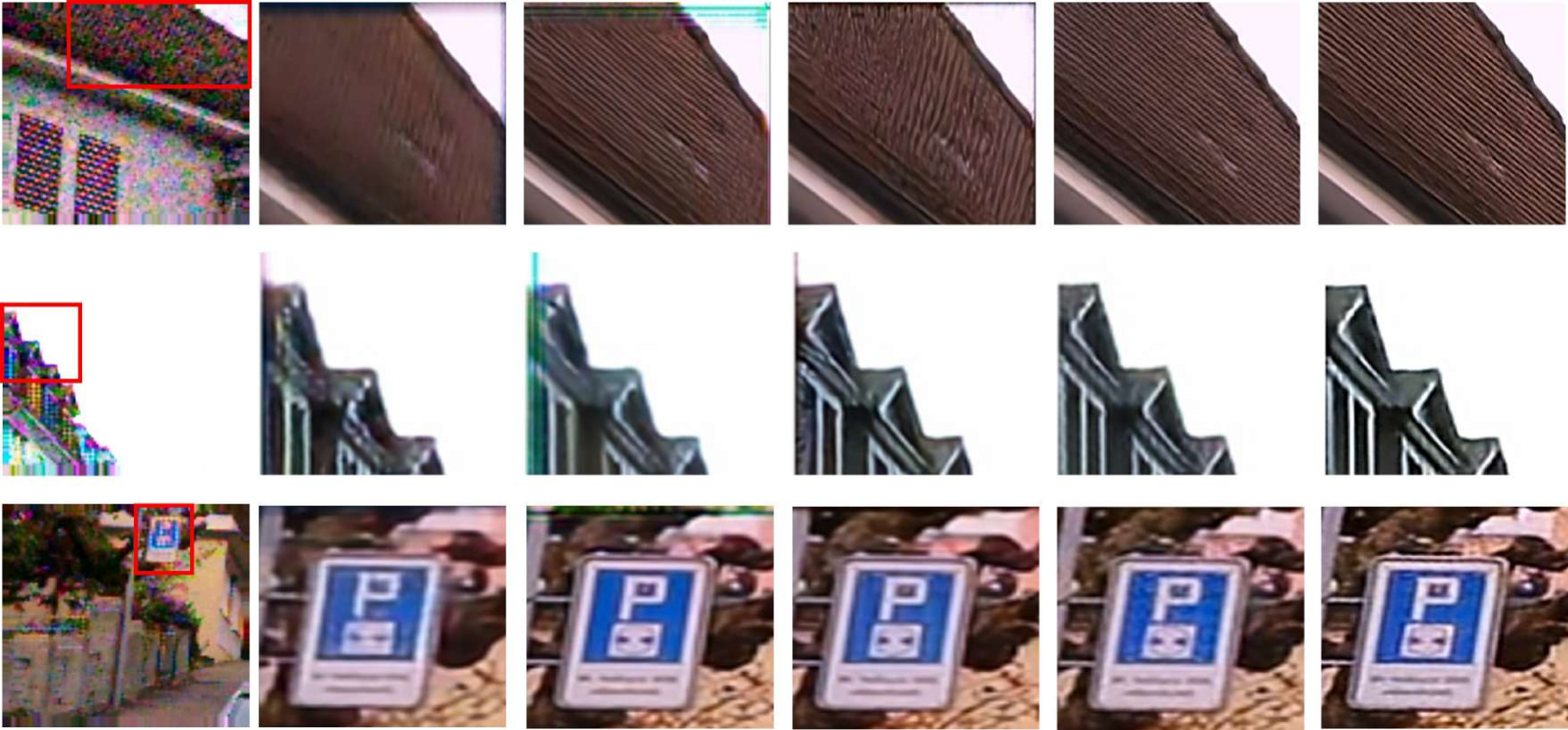}
    \scriptsize Base frame \qquad \qquad \qquad \qquad DBSR~\cite{bhat1} \qquad \qquad \qquad \qquad LKR~\cite{lecouat2021lucas} \qquad \qquad \qquad \qquad MFIR \cite{bhat2} \qquad \qquad \qquad \qquad \quad BIPNet \cite{dudhane} \qquad \qquad \qquad \qquad Ours
    \caption{Comparisons for $\times4$ burst SR on SyntheticBurst~\cite{bhat1}. The proposed approach produces more sharper and visually-faithful results than other competing approaches.}
    \label{fig: S_NTIRE_21_syn_val}
\end{figure*}

\begin{figure*}[t]
    \centering
    \includegraphics[width=1\linewidth]{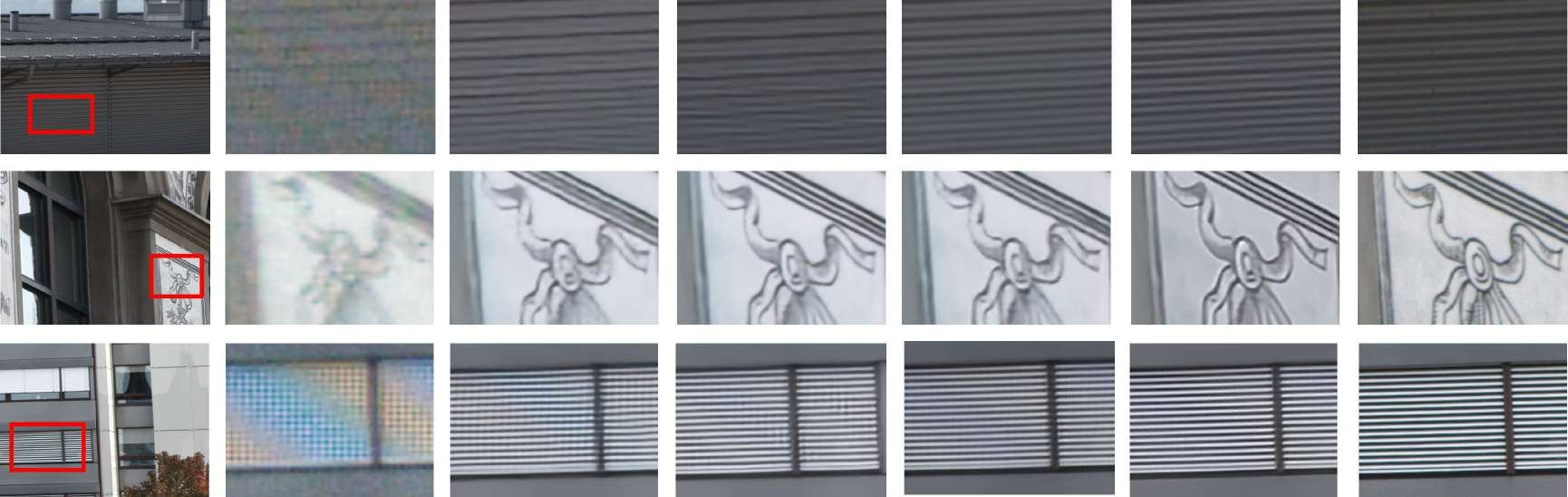}
    \scriptsize HR Image \qquad \qquad \qquad \qquad Base frame \qquad \qquad \qquad DBSR~\cite{bhat1} \qquad \qquad \qquad MFIR~\cite{bhat2} \qquad \qquad \qquad BIPNet \cite{dudhane} \qquad \qquad \qquad Ours \qquad \qquad \qquad Ground Truth
    \caption{Comparisons for $\times4$ burst super-resolution on Real BurstSR dataset~\cite{bhat1}. The proposed approach produces more sharper and cleaner results than other competing approaches.}
    \label{fig: S_NTIRE_21_real_val}
\end{figure*}

\begin{figure*}[t]
    \centering
    \includegraphics[width=1\linewidth]{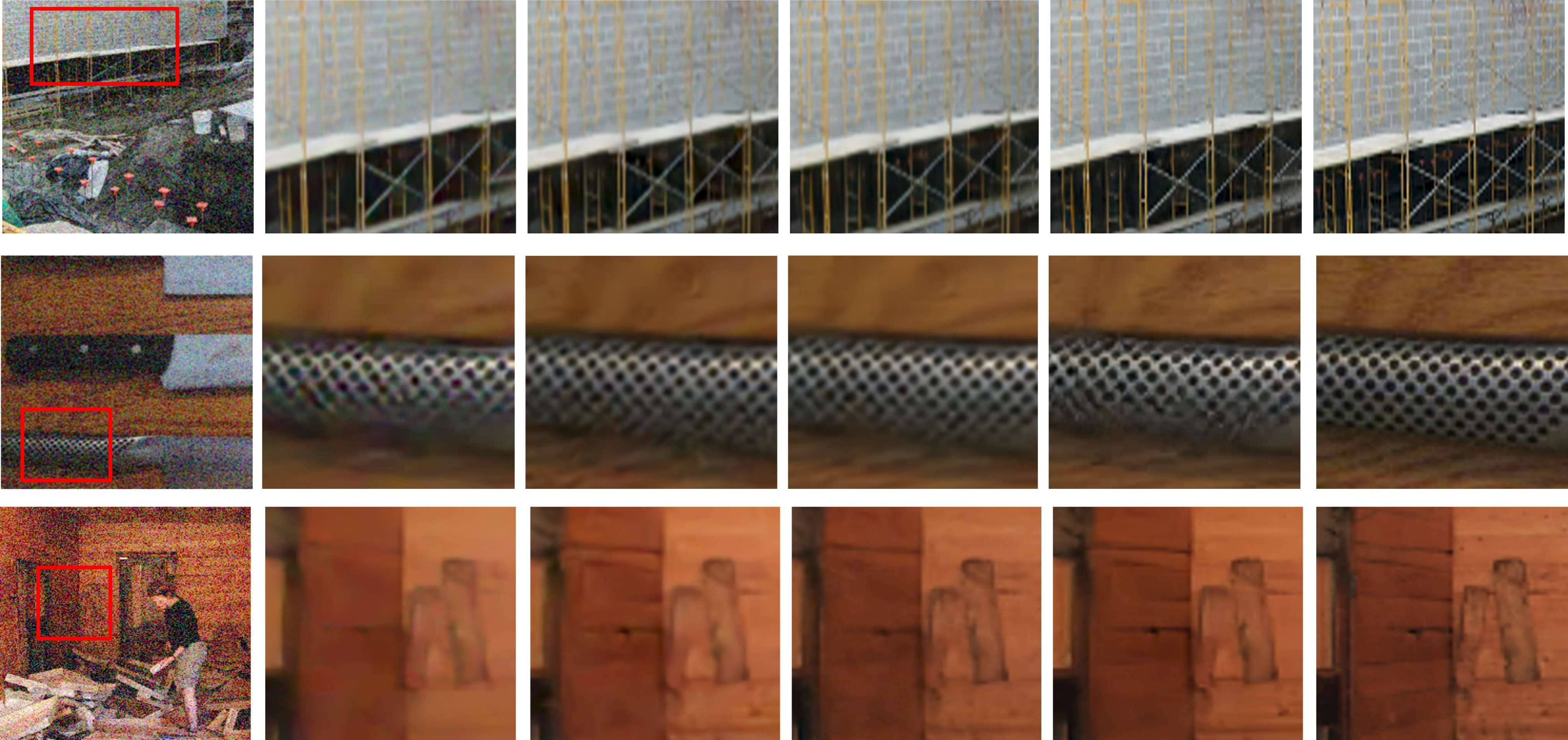}
    \scriptsize HR Image \qquad \qquad \qquad \qquad BPN \cite{gray} \qquad \qquad \qquad \qquad MFIR \cite{bhat2} \qquad \qquad \qquad \qquad BIPNet \cite{dudhane} \qquad \qquad \qquad \qquad \qquad Ours \qquad \qquad \qquad \qquad Ground Truth
    \caption{Comparisons for burst de-noising on color dataset~\cite{color}. The image reproduction quality of our proposed GMTNet is more faithful to the ground-truth than other competing approaches.}
    \label{fig: S_color_denoising}
\end{figure*}

\begin{figure*}[t]
    \centering
    \includegraphics[width=1\linewidth]{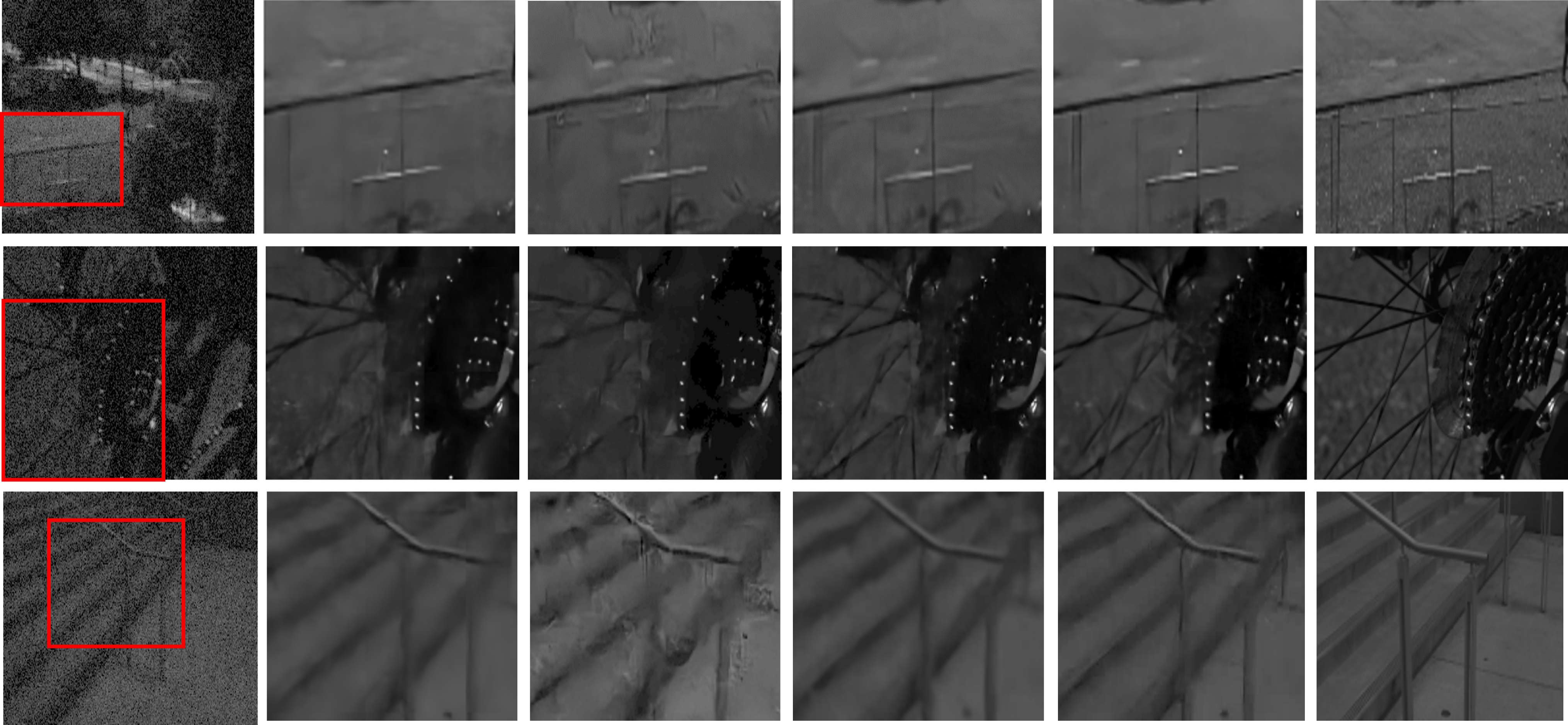}
    \scriptsize HR Image \qquad \qquad \qquad \qquad BPN \cite{gray} \qquad \qquad \qquad \qquad MFIR \cite{bhat2} \qquad \qquad \qquad \qquad BIPNet \cite{dudhane} \qquad \qquad \qquad \qquad \qquad Ours \qquad \qquad \qquad \qquad Ground Truth
    \caption{Comparisons for burst de-noising on gray-scale dataset~\cite{gray}. Our proposed GMTNet preserve the fine image details.}
    \label{fig: S_gray_denoising}
\end{figure*}

\begin{figure*}[t]
    \centering
    \includegraphics[width=1\linewidth]{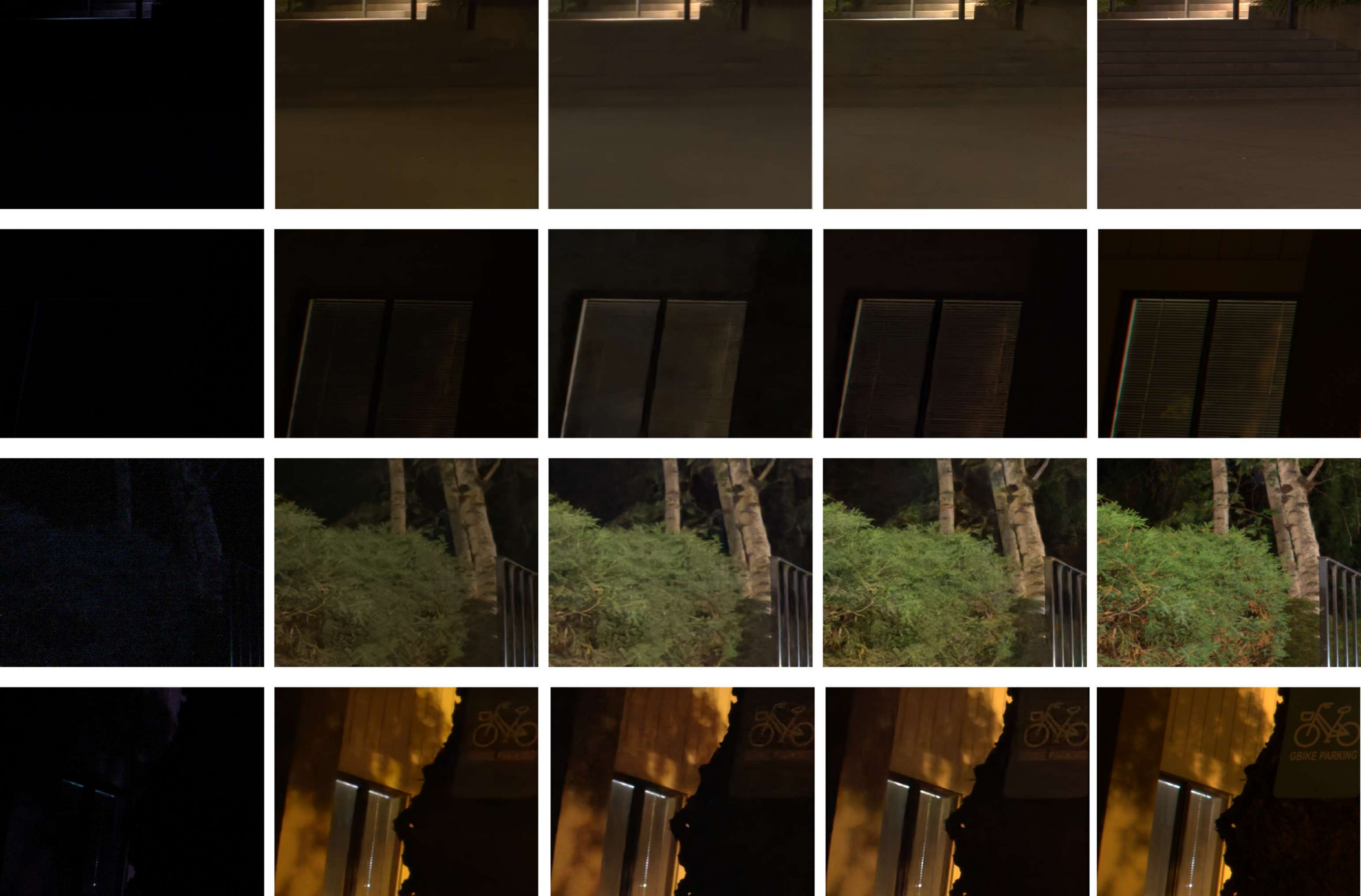}
     \scriptsize RAW input  \qquad \qquad \qquad \qquad \qquad \qquad  LLED \cite{karadeniz} \qquad \qquad \qquad \qquad \qquad \qquad  BIPNet \cite{dudhane} \qquad \qquad \qquad \qquad \qquad \qquad Ours \qquad \qquad \qquad \qquad \qquad \qquad GT image 
    \caption{Comparisons for burst low-light enhancement on SONY-subset~\cite{learning}. GMTNet generates sharper result with structural fidelity.}
    \label{fig: S_low-light}
\end{figure*}

We present more images reconstructed by our GMTNet and those of the other competing approaches as qualitative examples for all the considered tasks.
The results demonstrated in Figure \ref{fig: S_NTIRE_21_syn_val} and Figure \ref{fig: S_NTIRE_21_real_val} clearly show the true potential of our method in successfully recovering fine-grained details from extremely challenging LR images in $\times$4 burst SR task for synthetic and real images, respectively. 
Additionally, results in Figure \ref{fig: S_color_denoising} and Figure \ref{fig: S_gray_denoising} show that our method performs favorably well on both color \cite{color} and gray-scale \cite{gray} noisy images. 
Particularly, it generates output more closer to the ground-truth compared to the existing SoTA approaches. 
Further, we provide more visual comparisons for burst low-light enhancement in Figure \ref{fig: S_low-light} to show the effectiveness of our model.

 \begin{figure*}[t]
            \centering
            \includegraphics[width=1\linewidth]{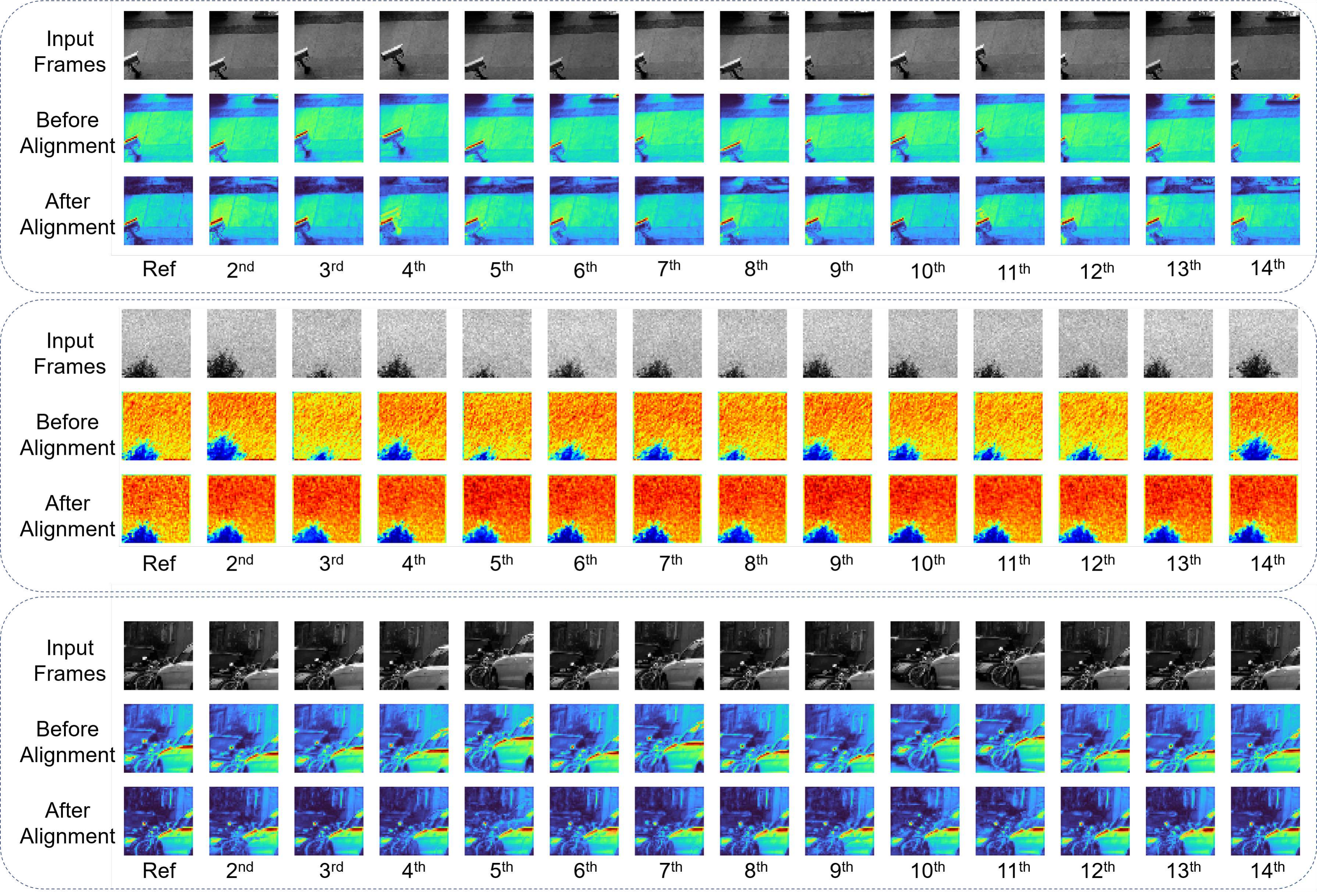}
             \scriptsize
            \caption{Feature map visualization before and after the proposed multi-scale burst feature alignment (MBFA) module on sample burst images from Synthetic BurstSR dataset. It is clearly observed that the proposed MBFA module aligns the burst frame features implicitly with respect to the reference frame features. \textbf{Highlighting point} is that the proposed MBFA module is implicitly trained along with the other modules of proposed network. We have not followed any separate supervision to train the MBFA module for alignment task.}
            \label{fig: feat_align1}
        \end{figure*}
 \begin{figure*}[t]
            \centering
            \includegraphics[width=1\linewidth]{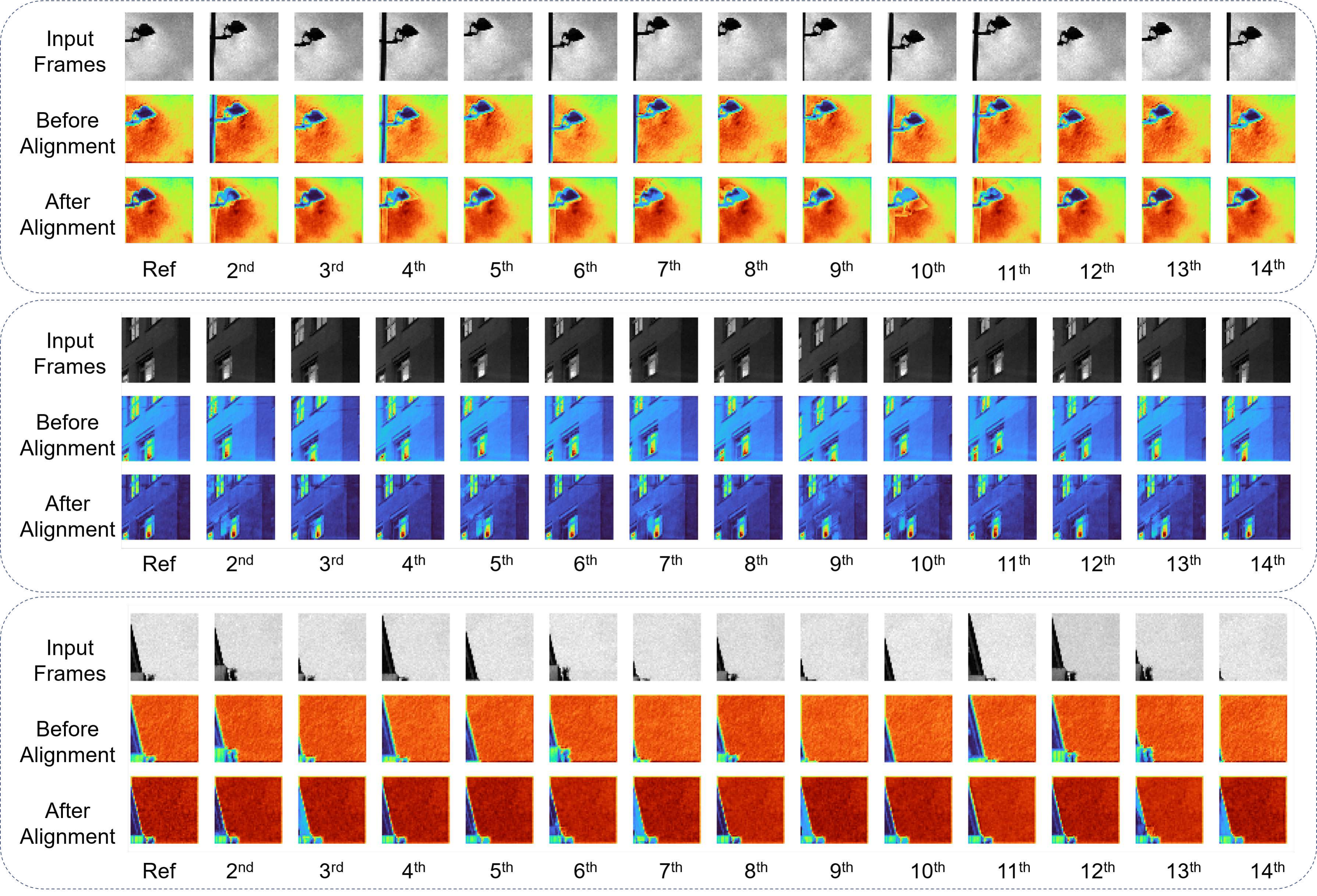}
             \scriptsize
            \caption{Feature map visualization before and after the proposed multi-scale burst feature alignment (MBFA) module on sample burst images from Synthetic BurstSR dataset. It is clearly observed that the proposed MBFA module aligns the burst frame features implicitly with respect to the reference frame features. \textbf{Highlighting point} is that the proposed MBFA module is implicitly trained along with the other modules of proposed network. We have not followed any separate supervision to train the MBFA module for the alignment task.}
            \label{fig: feat_align2}
        \end{figure*}

 \begin{figure*}[t]
            \centering
            \includegraphics[width=0.9\linewidth]{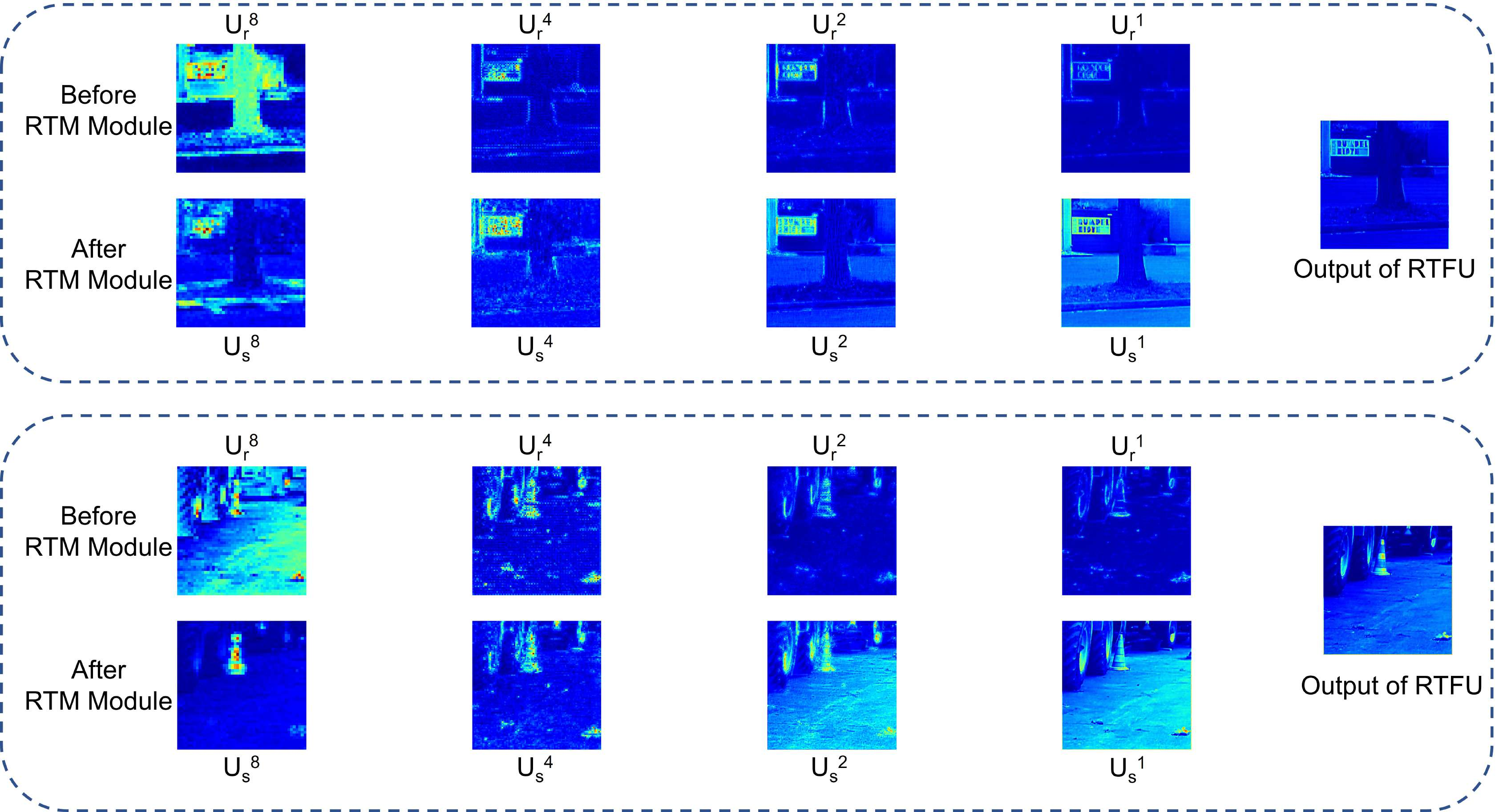}
             \scriptsize
            \caption{Feature map visualization before and after the proposed resolution transfer merging (RTM) module on sample burst images from Synthetic BurstSR dataset. It is clearly observed that for any pair of $U_r$ and $U_s$, latter is having more details than former one. The reason behind this is every $U_s$ feature response is obtained by fusion of all the $U_r$ feature responses. Thus, our RTM module benefits from the efficient extraction of features in both low and high resolution space and makes more effective use of information to complete the restoration of sharp regions.}
            \label{fig: feat_MM}
        \end{figure*}
\section{Feature Map Visualization}
\label{visualization}
Here, we have visualized the  feature maps of all the (14) frames before and after applying the proposed alignment (MBFA) (Fig. \ref{fig: feat_align1} and \ref{fig: feat_align2}). 
We note that all the unaligned input frames get aligned well with the reference frame, thereby demonstrating the effectiveness of the proposed MBFA module.
Similarly, in Figure \ref{fig: feat_MM}, we provide feature map visualizations before and after applying the resolution transfer merging (RTM) module of the proposed up-sampler (RTFU). 
It can be seen that for any $U_r$ and $U_s$ pair, the $U_s$ maps have more details than the $U_r$ maps.
Our RTM module benefits from the efficient feature extraction in both low and high-resolution spaces. 
Thus, we can conclude that the proposed modules enhances the feature representations as well as performs the assigned task dedicatedly.

\section{Future Work} 
\label{future}
While GMTNet emerges as a strong backbone architecture across several benchmarks, the proposed modules are extensible and can be well transferred to other burst image restoration applications including de-blurring and satellite imaging. 
Also, the proposed modules are suitable for majority of video restoration tasks.
\end{document}